\documentclass{article}

\makeatletter
\def\input@path{{iclr2027/}}
\makeatother
\usepackage[T1]{fontenc}
\usepackage{iclr2027_conference,times}
\iclrfinalcopy
\usepackage{amsmath,amssymb}
\usepackage{graphicx}
\graphicspath{{figures/}{iclr2027/figures/}}
\usepackage{booktabs}
\usepackage{tikz}
\usetikzlibrary{arrows.meta,positioning}
\usepackage{array}
\usepackage{colortbl}
\usepackage{placeins}
\usepackage{wrapfig}
\usepackage{multirow}
\newsavebox{\guatablebox}
\makeatletter
\newcommand{\finishwrap}{\par
  \ifnum\c@WF@wrappedlines>1
    \dimen@=\baselineskip
    \multiply\dimen@ by \numexpr\c@WF@wrappedlines-1\relax
    \vskip\dimen@
    \global\c@WF@wrappedlines=0
    \WF@finale
  \fi}
\makeatother

\newcommand{\dinomark}{\smash{\textsuperscript{*}}}
\usepackage{hyperref}
\usepackage{url}
\hypersetup{hidelinks}
\definecolor{basegray}{RGB}{233,233,233}
\definecolor{timegreen}{RGB}{212,244,214}
\definecolor{weightblue}{RGB}{211,233,248}
\definecolor{teacherpink}{RGB}{250,216,218}
\definecolor{lengthpurple}{RGB}{230,220,249}
\definecolor{batchyellow}{RGB}{251,237,203}
\colorlet{oursorange}{batchyellow}
\newcommand{\gvae}{GuaVAE}

\newcommand{\E}{\mathbb{E}}
\newcommand{\Lalign}{\mathcal{L}_{\mathrm{align}}}
\newcommand{\Ldiff}{\mathcal{L}_{\mathrm{diff}}}

\title{\centering
  GuaVAE: Guidance-Aligned End-to-End\\
  VAE Training for Latent Diffusion}

\author{%
  \makebox[\dimexpr\textwidth-2\tabcolsep\relax][s]{%
    \begin{tabular}[t]{@{}c@{}}
      \textbf{Andrey Chugunov}\\[0.35em]
      \normalfont\small HSE University
    \end{tabular}\hfill
    \begin{tabular}[t]{@{}c@{}}
      \textbf{Maksim Nakhodnov}\\[0.35em]
      \normalfont\small MSU\\
      \normalfont\small Constructor University
    \end{tabular}\hfill
    \begin{tabular}[t]{@{}c@{}}
      \textbf{Aibek Alanov}\\[0.35em]
      \normalfont\small HSE University
    \end{tabular}%
  }%
}

\begin{document}
\raggedbottom
\maketitle
\lhead{}
\renewcommand{\headrulewidth}{0pt}
\begin{abstract}
Recent studies have demonstrated that pretrained visual encoders such
as DINO can help to improve latent spaces for diffusion modeling.
However, these approaches depend on an external representation model.
We investigate whether diffusion transformers (DiTs) themselves can
provide supervision for learning better latent spaces through VAE tuning.
Prior work has shown that classifier-free guidance (CFG) can enhance
internal DiT features, making them effective targets for representation
alignment during training. We introduce GuaVAE, an end-to-end VAE tuning method that extends
this objective to the VAE encoder.
To encourage the VAE to learn a more generation-friendly latent space,
we align a trainable DiT's features with CFG-enhanced features obtained
at low noise levels from a frozen pretrained DiT.
We backpropagate the alignment loss through the trainable DiT
to the VAE encoder. Experiments show that
the tuned VAE improves generation quality across DiT backbone sizes,
achieving performance competitive with DINO-supervised VAEs.
We also observe better generation quality when using VAEs tuned with
larger DiTs, supporting diffusion representations as a scalable source
of supervision. A DiT trained with the tuned VAE and CFG-enhanced feature alignment
achieves a gFID of $3.73$ on ImageNet $256 \times 256$ without CFG,
outperforming self-contained representation alignment baselines
as well as DINO-guided REPA. With CFG, it achieves a competitive
gFID of $1.43$.
\end{abstract}

\suppressfloats[t]
\section{Introduction}
\label{sec:introduction}
\begin{figure}[t]
\centering
\includegraphics[width=\linewidth]{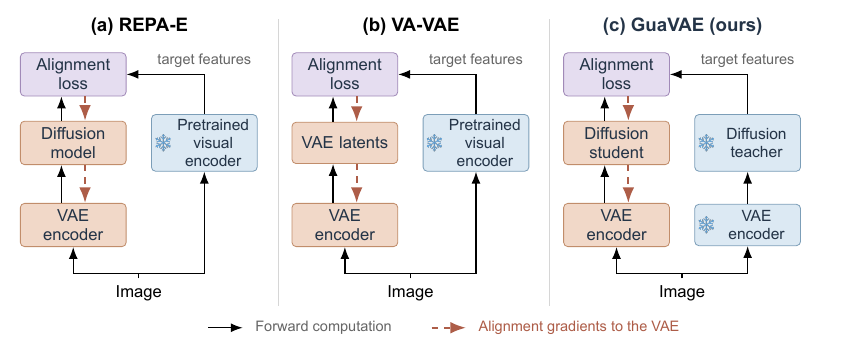}
\vspace{-2em}
\caption{\textbf{Representation supervision for improving VAE latent spaces.}
(a) REPA-E aligns diffusion features with a pretrained visual encoder and
backpropagates the alignment loss through the diffusion model to the VAE.
(b) VA-VAE aligns VAE latents with a pretrained visual encoder.
(c) \gvae{} replaces the external visual encoder with a diffusion teacher.
Snowflakes denote frozen modules; dashed arrows show alignment gradients
to the VAE encoder. Projection heads, noising, feature guidance,
and other objectives are omitted; Figure~\ref{fig:method}
details the diffusion alignment pathway.}
\label{fig:intro_comparison}
\vspace{-0.5em}
\end{figure}

Diffusion models have become a widely used approach to high-quality image
generation \citep{ho2020ddpm,song2021sde,dhariwal2021diffusion}.
To reduce the cost of modeling high-resolution images, latent diffusion
models operate in a compressed latent space produced by an autoencoder,
typically a variational autoencoder (VAE) \citep{rombach2022ldm}.
The encoder maps images to latent representations, the diffusion model learns
to generate samples in this latent space, and the decoder maps those
samples back to images.
The choice of latent space affects the quality of generated images:
some latent distributions are easier for a diffusion model to learn than
others. This ease of learning is referred to as \emph{diffusability}
\citep{skorokhodov2025diffusability}.

Representation alignment offers one way to improve VAE latent spaces.
VA-VAE (Figure~\ref{fig:intro_comparison}b) aligns VAE latents with features
from a pretrained visual encoder such as DINO \citep{yao2025vavae}.
REPA-E (Figure~\ref{fig:intro_comparison}a) instead aligns intermediate
diffusion features with the visual encoder's features and backpropagates
the alignment loss through the diffusion model to the VAE encoder
\citep{leng2025repae}. Both approaches improve generation with the resulting
VAE, but require an external pretrained encoder to provide the alignment
targets.
These encoders are trained with objectives that are not designed
 to facilitate generation. RePR argues that recognition-oriented
 representation learning can suppress information important for generative
 modeling \citep{baade2026repr}.

Representation alignment also improves diffusion training itself.
REPA accelerates the training of diffusion transformers (DiTs)
by aligning their intermediate features with DINO features
\citep{yu2025repa}.
However, Self-Flow reports that stronger external encoders do not
 necessarily improve generation and that external alignment yields
 diminishing returns as the generative model grows
 \citep{chefer2026selfflow}. Together, these findings motivate investigating
 alternative training signals for VAE tuning.

Self-contained alternatives such as SRA and LayerSync
guide earlier diffusion layers with representations from deeper layers,
avoiding this external dependency
\citep{jiang2025sra,haghighi2025layersync}.
Self-Transcendence further improves the quality of internal targets
\citep{sun2026selftranscendence}. It first trains a diffusion teacher
using an auxiliary VAE-based objective to improve its early-layer
representations. It then applies classifier-free guidance (CFG)
to the teacher's intermediate features to strengthen their class-specific
information. Aligning a new diffusion model with these features achieves
generation quality competitive with DINO-guided REPA.
External features can
improve VAE latent spaces, and internal diffusion features can replace
external features in diffusion training. This motivates our central
question: \textbf{Can internal diffusion representations be used to improve
a VAE's latent space?}

We introduce \gvae{} (\textbf{Gu}idance-\textbf{A}ligned \textbf{VAE}), an end-to-end method
for tuning a VAE using intermediate features from a frozen DiT teacher.
To strengthen the semantic signal, we apply CFG to the teacher's features to obtain the alignment targets.
We fine-tune the VAE and pass its latents to a trainable DiT student, which aligns its own intermediate features with these targets.
Backpropagating the alignment loss through the student updates the
VAE encoder, adapting its latent space for downstream generation
(Figure~\ref{fig:intro_comparison}c).
The teacher's noise level is important: independently sampling its timestep from a low-noise interval yields substantially better VAEs than the naive approach of matching teacher and student timesteps.

An additional question is whether stronger diffusion models can produce
better VAEs. \space In our experiments, increasing the
teacher and student backbones together improves downstream generation
across the tested model sizes, with the evaluation model held fixed.
The resulting VAEs narrow the gap in generation quality with
DINO-supervised VAEs,
suggesting that increasing diffusion-model capacity can improve the
effectiveness of this supervision.

Our contributions are summarized as follows:
\begin{itemize}
\item We introduce \gvae{}, a method for tuning VAE latent spaces using
internal diffusion representations without an external visual encoder.
The resulting VAEs improve generation with DiTs and narrow the gap
in generation quality with DINO-supervised VAEs as the DiT backbone
used for VAE tuning increases in size.
\item We combine \gvae{} with Self-Transcendence to obtain a self-contained
generative method.
On ImageNet $256\times256$, it achieves gFID $3.73$ without sampling CFG,
outperforming self-contained representation-alignment methods and DINO-guided
REPA, and a competitive gFID of $1.43$ with sampling CFG.
\end{itemize}

\section{Related Work}
\label{sec:related_work}
\textbf{Visual tokenizers and latent diffusion.}
Visual tokenizers map images to compact representations for generative
modeling. VAEs learn continuous latent variables through variational
inference \citep{kingma2014vae}, while VQ-VAE learns a discrete codebook
\citep{oord2017vqvae}. VQGAN combines quantization with perceptual and
adversarial objectives to preserve visual detail under compression
\citep{esser2021vqgan}.
Diffusion and score-based models generate samples by reversing a
noise-corruption process \citep{ho2020ddpm,song2021sde,dhariwal2021diffusion}.
LSGM and latent diffusion move this modeling problem to an autoencoder's
latent space, reducing the cost of high-resolution generation
\citep{vahdat2021lsgm,rombach2022ldm}. Transformer backbones such as
U-ViT and DiT provide scalable architectures for diffusion modeling
\citep{bao2023uvit,peebles2023dit}. Flow matching learns transport vector
fields \citep{lipman2023flow}, and SiT studies flow and diffusion within
a scalable interpolant framework \citep{ma2024sit}.

\textbf{Representation learning for diffusion.}
Auxiliary representation objectives can improve diffusion training.
MaskDiT combines denoising with masked-patch reconstruction
\citep{zheng2024maskdit}, while SD-DiT adds self-supervised discriminative
learning \citep{zhu2024sddit}. REPA directly aligns intermediate diffusion
features with representations from pretrained visual encoders
\citep{yu2025repa}, including DINOv2 \citep{oquab2024dinov2}.
iREPA studies which properties make these targets effective and
emphasizes spatial structure over global semantic information
\citep{singh2025irepa}.
Other methods obtain supervision from the generative system itself.
SRA and LayerSync align internal diffusion representations
\citep{jiang2025sra,haghighi2025layersync}; SRA2 uses VAE representations
\citep{wang2026sra2}. Self-Transcendence combines VAE-based teacher
preparation with guided diffusion features to provide effective
alignment targets without a separate visual encoder
\citep{sun2026selftranscendence}.

\textbf{Autoencoders for downstream generation.}
Recent work studies latent-space properties that affect generation.
EQ-VAE regularizes spatial equivariance \citep{kouzelis2025eqvae}, while
\citet{skorokhodov2025diffusability} analyze latent frequency structure
and introduce scale-equivariance regularization. A broader study of
latent diffusability examines spatial structure and semantic
organization as predictors of generation quality
\citep{zhong2026diffusability}.
MAETok uses masked autoencoding to learn representations suitable for
diffusion \citep{chen2025maetok}. Representation autoencoders pair
pretrained representation encoders with learned decoders and adapt
diffusion to their high-dimensional latents \citep{zheng2025rae}.
VA-VAE instead aligns VAE latents with pretrained visual features
\citep{yao2025vavae}, and REPA-E propagates representation alignment
through the diffusion model to the VAE encoder \citep{leng2025repae}.
Diffusion as Self-Distillation (DSD) addresses latent collapse through
a self-distillation formulation, sharing a transformer backbone across
encoding, reconstruction, and diffusion \citep{wang2025dsd}.
Concurrent work GenFirst stabilizes joint training through
posterior-entropy regularization and a schedule that prioritizes
generation before progressively strengthening reconstruction
\citep{zheng2026genfirst}.

\section{Method}
\label{sec:method}
\begin{figure}[!t]
\centering
\includegraphics[width=\linewidth]{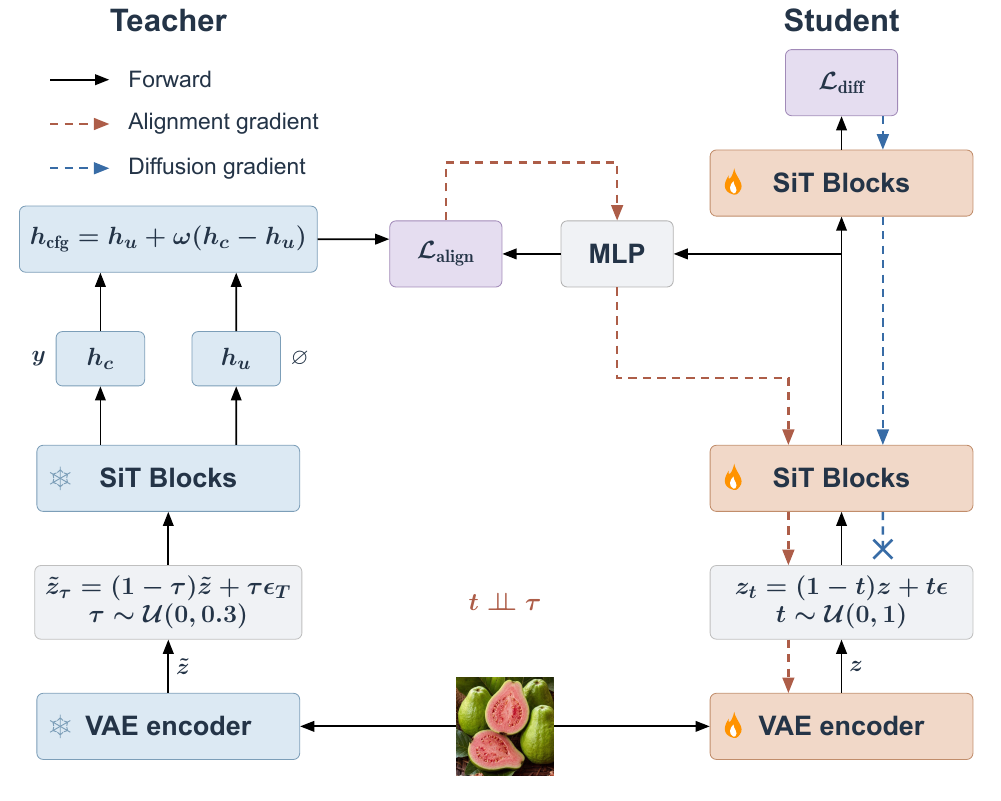}
\vspace{-2em}
\caption{\textbf{\gvae{} tuning.} The prepared teacher supplies CFG-enhanced
targets at independently sampled low-noise timesteps. Red dashed arrows
trace alignment gradients; blue dashed arrows trace diffusion gradients.
The cross marks the diffusion-gradient boundary; snowflakes and flames
denote frozen and trainable modules. VAE and student updates alternate
(Section~\ref{sec:e2e}); normalization and reconstruction objectives are omitted.}
\vspace{-1em}
\label{fig:method}
\end{figure}

Our training system consists of a trainable VAE and diffusion student,
and a diffusion teacher with its own VAE (Figure~\ref{fig:method}).
We prepare the teacher following the first stage of
Self-Transcendence \citep{sun2026selftranscendence}: an auxiliary head
predicts the diffusion velocity from an early layer, alongside the final
diffusion prediction. Appendix~\ref{app:training_details} specifies this
objective. This preparation improves downstream generation over a
teacher trained only with the diffusion loss
(Table~\ref{tab:ablations}(d)).
During end-to-end tuning, the teacher
and its VAE remain frozen. The alignment loss updates the student and
reaches the trainable VAE encoder through it.

\subsection{Latent Diffusion Setup}
\label{sec:preliminaries}
Let $E$ and $D$ denote the trainable VAE encoder and decoder, and $f$ the
diffusion student. For an image $x$ with class label $y$, we sample
$u\sim q_E(u\mid x)$ from the encoder posterior and normalize it to
$z=B(u)$. Following REPA-E \citep{leng2025repae}, $B$ is a non-affine
batch-normalization layer that tracks the changing latent statistics.
Following SiT \citep{ma2024sit}, we form the noisy latent
\begin{equation}
 z_t=(1-t)z+t\epsilon,\qquad
 t\sim\mathcal{U}(0,1),\quad \epsilon\sim\mathcal{N}(0,I).
 \label{eq:interpolant}
\end{equation}
The student predicts the velocity $\epsilon-z$ using the diffusion loss
\begin{equation}
 \Ldiff=\E\!\left[\frac{1}{N}\left\|
 f(z_t,t,y)-(\epsilon-z)\right\|_2^2\right],
 \label{eq:diffusion}
\end{equation}
where $N$ is the number of scalar latent entries per image and the norm
is taken over the flattened tensor. Following REPA-E, we detach the VAE
latents when computing this loss, so its gradients update only the
diffusion student. The alignment loss defined below provides the gradient
path through the student to the VAE encoder.

\subsection{Constructing Diffusion Alignment Targets}
\label{sec:guidance}
Let $\tilde z$ denote the normalized latent of the same image $x$ from
the teacher VAE, and let $f_T$ denote the diffusion teacher. We sample
its timestep $\tau$ independently of the student timestep $t$:
\begin{equation}
 \tilde z_\tau=(1-\tau)\tilde z+\tau\epsilon_T,
 \qquad \tau\sim\mathcal{U}(a,b).
 \label{eq:teacher_input}
\end{equation}
For same-shape SD-VAE tuning, $\epsilon_T=\epsilon$ from
Eq.~\eqref{eq:interpolant}. For cross-VAE tuning with different latent
shapes, the two Gaussian draws are independent
(Appendix~\ref{app:training_details}).

\textbf{Feature guidance.}
Following Self-Transcendence \citep{sun2026selftranscendence}, we enhance
the teacher's intermediate features using CFG. Conditional and
unconditional passes use the same teacher input:
\begin{equation}
 \begin{gathered}
 h_c = h_T(\tilde z_\tau,\tau,y), \qquad
 h_u = h_T(\tilde z_\tau,\tau,\varnothing),\\
 h_{\mathrm{cfg}} = h_u + \omega(h_c-h_u),
 \end{gathered}
 \label{eq:feature_cfg}
\end{equation}
where $h_T$ extracts features from a selected teacher layer and $\omega$
controls feature guidance. This CFG constructs training targets and is
separate from CFG applied during image sampling. Our ablation shows that conditional teacher features alone
($\omega=1$) do not improve downstream generation, whereas
CFG-enhanced targets do (Table~\ref{tab:ablations}(c)).

\textbf{Choosing the teacher timestep.}
Our aim is to obtain targets that improve the VAE as well as the student.
In a preliminary shared-timestep run, PCA visualizations of the VAE
latents show clearer object structure at $1{,}650$ iterations, but this
structure deteriorates by $100$K iterations despite continued alignment,
and the latents again resemble those of the teacher's VAE
(Appendix~\ref{app:shared_timestep}, Figure~\ref{fig:shared_timestep}). We hypothesize that, when teacher and student noise levels match, alignment can be achieved mainly through updates to the student diffusion model. This provides little incentive for the VAE to learn a latent space better suited for generation. We therefore construct targets whose noise level does not
track the student's timestep. Clean visual-encoder targets in REPA-E
\citep{leng2025repae} also motivate using less noisy targets.
We sample the teacher timestep independently from a low-noise interval,
retaining the full range for the student. Unlike SRA
\citep{jiang2025sra}, this does not require lower teacher noise for
every pair. Independently sampling the teacher timestep improves downstream
generation even when both timesteps follow the same full-range
distribution. Restricting the teacher to a low-noise interval provides
a further improvement (Table~\ref{tab:ablations}(b)), supporting both
components of our sampling strategy.

\subsection{Updating the VAE through the Student}
\label{sec:e2e}
Let $h(z_t,t,y)$ denote the student features and $P$ a trainable projection
head. We align the projected features with the CFG-enhanced teacher target:
\begin{equation}
 \Lalign=\E\!\left[\frac{1}{N}
 \left\|P(h(z_t,t,y))-h_{\mathrm{cfg}}\right\|_2^2\right].
 \label{eq:alignment}
\end{equation}
Here $N$ is the number of scalar feature entries per image, so the loss
averages over both tokens and channels. The norm is taken over flattened
features. The teacher target is fixed and receives no gradients.
Since $z_t$ depends on the trainable encoder, the alignment loss
backpropagates through $P$ and the student to the VAE encoder.

\textbf{VAE regularization.} To preserve reconstruction quality during tuning, we retain the VAE's
standard regularization losses, denoted collectively by
$\mathcal{L}_{\mathrm{reg}}$: pixel reconstruction, LPIPS, adversarial,
and KL terms, following REPA-E \citep{leng2025repae}.

\textbf{Training objectives and alternating updates.} The two updates minimize separate sums of losses:
\begin{equation}
\begin{aligned}
 \mathcal L_{\mathrm{VAE}} &= \mathcal L_{\mathrm{reg}}+\lambda_V\Lalign,\\
 \mathcal L_{\mathrm{student}} &= \Ldiff+\lambda_S\Lalign.
\end{aligned}
\label{eq:training_objectives}
\end{equation}
The VAE update optimizes $E$ and $D$, holding $f$ and $P$ fixed while
differentiating through their inputs. The student update optimizes $f$
and $P$ with detached VAE latents. The teacher remains frozen in both
updates. The weights $\lambda_V$ and $\lambda_S$ control alignment in
the respective objectives. Training details are in
Appendix~\ref{app:training_details}.

\subsection{Generation after VAE Tuning}
\label{sec:training_modes}
After tuning, we freeze the VAE and its latent normalization. To assess
the VAE independently of the jointly trained student, we train a new
diffusion model from scratch using only $\Ldiff$.

For our final generative model, we instead continue the existing student
with $\Ldiff$ alone. This retains the benefit of feature alignment during
the initial stage, while sampling uses the trained diffusion model,
inverse latent normalization, and the VAE decoder.
Section~\ref{sec:synergy} compares continuation with training a new
feature-aligned student on the tuned VAE.

\FloatBarrier

\section{Experiments}
\label{sec:experiments}
We evaluate whether diffusion representations improve VAE latent spaces for generation, how the gains depend on model size, and how to build a generator from the tuned VAE and student.

\begin{table}[!tbp]
\centering
\caption{\textbf{VAE quality for downstream generation.} gFID of new generators
trained on frozen VAEs. Both Ours VAEs use SiT-XL tuning with $\lambda_V=0.05$
and an SD-VAE teacher. $^{*}$: DINO supervision.}
\label{tab:vae_quality}
\small
\begin{minipage}[t]{0.315\linewidth}
\centering\textbf{(a) SiT-B, $f8d4$}\par\smallskip
\begingroup
\setlength{\tabcolsep}{3pt}
\def\guapanelrows{
\toprule VAE & gFID$\downarrow$\\\midrule
SD-VAE & 37.02\\
EQ-VAE & 31.20\\
\rowcolor{oursorange}Ours & 22.53\\
\midrule
REPA-E\dinomark & 20.50\\\bottomrule
}
\sbox{\guatablebox}{\begin{tabular}{lr}\guapanelrows\end{tabular}}
\setlength{\tabcolsep}{\dimexpr3pt+(\linewidth-\wd\guatablebox)/4\relax}
\begin{tabular}{lr}\guapanelrows\end{tabular}
\endgroup
\end{minipage}\hfill
\begin{minipage}[t]{0.315\linewidth}
\centering\textbf{(b) SiT-B, $f16d32$}\par\smallskip
\begingroup
\setlength{\tabcolsep}{3pt}
\def\guapanelrows{
\toprule VAE & gFID$\downarrow$\\\midrule
IN-VAE & 39.31\\
\rowcolor{oursorange}Ours & 19.30\\
\midrule
REPA-E\dinomark & 17.31\\
VA-VAE\dinomark & 13.85\\\bottomrule
}
\sbox{\guatablebox}{\begin{tabular}{lr}\guapanelrows\end{tabular}}
\setlength{\tabcolsep}{\dimexpr3pt+(\linewidth-\wd\guatablebox)/4\relax}
\begin{tabular}{lr}\guapanelrows\end{tabular}
\endgroup
\end{minipage}\hfill
\begin{minipage}[t]{0.315\linewidth}
\centering\textbf{(c) SiT-XL}\par\smallskip
\begingroup
\setlength{\tabcolsep}{3pt}
\def\guapanelrows{
\toprule Latent & VAE & gFID$\downarrow$\\\midrule
\multirow{2}{*}{$f8d4$} & SD-VAE & 17.46\\
& \cellcolor{oursorange}Ours & \cellcolor{oursorange}10.29\\
\midrule
\multirow{2}{*}{$f16d32$} & IN-VAE & 18.00\\
& \cellcolor{oursorange}Ours & \cellcolor{oursorange}8.13\\\bottomrule
}
\sbox{\guatablebox}{\begin{tabular}{clr}\guapanelrows\end{tabular}}
\setlength{\tabcolsep}{\dimexpr3pt+(\linewidth-\wd\guatablebox)/6\relax}
\begin{tabular}{clr}\guapanelrows\end{tabular}
\endgroup
\end{minipage}
\end{table}

\textbf{Setup.}
\label{sec:setup}
We evaluate class-conditional generation on ImageNet at $256\times256$
resolution \citep{deng2009imagenet}. We tune SD-VAE ($f8d4$)
\citep{rombach2022ldm} and IN-VAE ($f16d32$), the ImageNet-trained VAE
provided by REPA-E \citep{leng2025repae}. Both are baseline VAEs trained
without DINO supervision. Here $f$ denotes spatial downsampling and
$d$ the number of latent channels. In the ImageNet frozen-VAE comparisons, we train a new SiT on each fixed VAE
for $400$K iterations using only the diffusion loss. Scaling increases teacher and student sizes together, from SiT-B to SiT-2XL, while holding the downstream generator fixed.

Our default schedule comprises $200$K iterations of teacher preparation
and $100$K of VAE tuning, with teacher timesteps sampled from $(0,0.3)$.
Feature guidance $\omega=30$ and student alignment $\lambda_S=0.5$
follow Self-Transcendence \citep{sun2026selftranscendence}. For the main runs, the VAE alignment weight is $\lambda_V=0.25$
for SiT-B and $0.05$ for SiT-XL. The smaller XL weight better preserves reconstruction (Appendix~\ref{app:reconstruction}). The scaling
comparison instead holds $\lambda_V=0.25$ fixed across model sizes. Full configurations are given in Appendix~\ref{app:training_details}.
We evaluate ImageNet generation using $50$K samples with the ADM evaluator
\citep{dhariwal2021diffusion} and $250$-step Euler--Maruyama SDE sampling.
Our sampling uses uniformly random class labels and no CFG except in
the guided comparison. Differences in published sampling protocols are described in
Appendix~\ref{app:evaluation}.
In the tables, $^{*}$ marks methods using DINO supervision.
Baseline sources and evaluation details are given in
Appendix~\ref{app:evaluation}.

\subsection{Improving VAE Quality for Downstream Generation}
\label{sec:fixed_vae}
\noindent\begin{minipage}[t]{0.49\linewidth}
\vspace{0pt}
Table~\ref{tab:vae_quality} evaluates the tuned VAEs with newly trained
diffusion models. For SD-VAE, tuning lowers gFID from $37.02$ to $22.53$
with SiT-B and from $17.46$ to $10.29$ with SiT-XL. Because both
generators are trained from scratch, these gains demonstrate the value
of the resulting latent space independently of the jointly trained student.
Our SD-VAE also improves on EQ-VAE, while DINO-supervised REPA-E remains
stronger in the SiT-B comparison.
Appendix~\ref{app:frozen_vae_cfg} reports results with CFG
for the same SiT-B models evaluated in Table~\ref{tab:vae_quality}(a).
\end{minipage}\hfill
\begin{minipage}[t]{0.49\linewidth}
\vspace{0pt}
\def\guascale{%
\toprule
Method & SiT & gFID$\downarrow$\\
\midrule
\rowcolor{oursorange}Ours & B & 23.88\\
\rowcolor{oursorange}Ours & L & 22.87\\
\rowcolor{oursorange}Ours & XL & 20.89\\
\rowcolor{oursorange}Ours & 2XL & \textbf{20.59}\\
\midrule
REPA-E\dinomark & B & \textbf{19.40}\\
REPA-E\dinomark & L & 19.85\\
REPA-E\dinomark & XL & 20.31\\
REPA-E\dinomark & 2XL & 20.02\\
\bottomrule
}
\sbox{\guatablebox}{\small\setlength{\tabcolsep}{5pt}\begin{tabular}{llr}\guascale\end{tabular}}
\begingroup
\makeatletter\def\@captype{table}\makeatother
\centering
\setlength{\abovecaptionskip}{0pt}
\caption{\textbf{SD-VAE scaling.} SiT denotes the tuning backbone. Evaluation uses a new SiT-B/2. Ours holds $\lambda_V=0.25$ fixed.}
\label{tab:vae_scale}
\small\setlength{\tabcolsep}{\dimexpr5pt+(\linewidth-\wd\guatablebox)/6\relax}
\begin{tabular}{llr}\guascale\end{tabular}\par
\endgroup

\end{minipage}\par

\textbf{Generalization to IN-VAE.}
The method also improves IN-VAE, whose latents have greater spatial
compression and more channels. To choose its teacher VAE, we compare
SD-VAE and IN-VAE teachers under identical SiT-B tuning settings:
SD-VAE gives gFID $21.06$, versus $26.12$ with IN-VAE. We therefore use
SD-VAE as the teacher VAE for both latent formats
(Appendix~\ref{app:teacher_vae}).

With this teacher, our selected XL-tuned IN-VAE lowers downstream gFID
from $39.31$ to $19.30$ \space with SiT-B and from $18.00$ to $8.13$
with SiT-XL. A gap to REPA-E and VA-VAE remains \space in the SiT-B comparison, but the
improvement shows that diffusion supervision is useful beyond SD-VAE.
Both Ours checkpoints in Table~\ref{tab:vae_quality} use
$\lambda_V=0.05$; reconstruction results for both latent formats are reported in
Appendix~\ref{app:reconstruction}.

\subsection{Scaling the Models Used for VAE Tuning}
\label{sec:scaling}
We next scale the teacher and student backbones used to tune SD-VAE,
keeping the alignment weight fixed, the batch size at $256$, and the
downstream architecture fixed at SiT-B/2. For each checkpoint, we
report mean gFID over three sampling seeds, with $50$K images generated
per seed. Mean gFID decreases consistently (Table~\ref{tab:vae_scale})
from $23.88$ with SiT-B to $20.59$ with SiT-2XL, approaching REPA-E
($19.40$).

REPA-E does not exhibit the same monotonic trend in this frozen-VAE
comparison. One possible explanation is the source of
supervision: REPA-E uses the same DINO encoder across these runs,
whereas our teacher grows together with the student. Increasing model
size therefore also increases the capacity of the model supplying our
alignment targets, which may make the supervision more effective.
Separately, REPA and REPA-E report limited gains from increasing the
DINO encoder size \citep{yu2025repa,leng2025repae}.
\finishwrap

\subsection{Component Analysis}
\label{sec:ablations}

\definecolor{guidanceorange}{HTML}{FFD9B3}

\begin{table}[!t]
\centering
\caption{\textbf{Effect of VAE tuning settings on downstream generation.} Color marks the varied setting. $\omega$: feature-guidance weight for teacher targets; $\omega=1$ uses conditional teacher features without guidance. VA: auxiliary velocity supervision; Std.: diffusion-only teacher. Iter. and BS refer to VAE tuning. PSNR/rFID: reconstruction by the tuned VAE.}
\label{tab:ablations}
\small
\def\guashrink{80}%
\def\guarows{\toprule
$\lambda_V$ & $\tau$ & $\omega$ & Teacher & Iter. & BS & PSNR$\uparrow$ & rFID$\downarrow$ & gFID$\downarrow$ & IS$\uparrow$\\
\midrule
\multicolumn{6}{l}{Original SD-VAE (no tuning)} & 26.91 & 0.71 & 37.02 & 40.38\\
\midrule
\multicolumn{10}{l}{\textit{(a) Alignment weight}}\\
\cellcolor{weightblue}0.1 & $(0,0.1)$ & 30 & VA & 100K & 32 & \textbf{25.42} & \textbf{0.85} & 26.14 & 54.51\\
\cellcolor{weightblue}0.25 & $(0,0.1)$ & 30 & VA & 100K & 32 & 25.13 & 1.00 & 25.15 & 55.83\\
\cellcolor{weightblue}0.5 & $(0,0.1)$ & 30 & VA & 100K & 32 & 24.66 & 1.05 & 24.69 & \textbf{56.79}\\
\cellcolor{weightblue}1 & $(0,0.1)$ & 30 & VA & 100K & 32 & 24.37 & 1.11 & \textbf{24.27} & 56.53\\
\cellcolor{weightblue}2 & $(0,0.1)$ & 30 & VA & 100K & 32 & 23.93 & 1.32 & 25.08 & 55.42\\
\midrule
\multicolumn{10}{l}{\textit{(b) Teacher timestep}}\\
0.25 & \cellcolor{timegreen}$\tau=t$ & 30 & VA & 100K & 32 & -- & -- & 35.02 & 41.00\\
0.25 & \cellcolor{timegreen}$(0,1.0)$ & 30 & VA & 100K & 32 & -- & -- & 27.68 & 51.25\\
0.25 & \cellcolor{timegreen}$(0,0.05)$ & 30 & VA & 100K & 32 & -- & -- & 25.52 & 54.86\\
0.25 & \cellcolor{timegreen}$(0,0.1)$ & 30 & VA & 100K & 32 & -- & -- & 25.15 & 55.83\\
0.25 & \cellcolor{timegreen}$(0,0.2)$ & 30 & VA & 100K & 32 & -- & -- & 24.82 & 55.72\\
0.25 & \cellcolor{timegreen}$(0,0.3)$ & 30 & VA & 100K & 32 & -- & -- & \textbf{23.89} & \textbf{57.97}\\
0.25 & \cellcolor{timegreen}$(0,0.4)$ & 30 & VA & 100K & 32 & -- & -- & 24.65 & 56.91\\
0.25 & \cellcolor{timegreen}$(0.1,0.3)$ & 30 & VA & 100K & 32 & -- & -- & 24.86 & 56.65\\
\midrule
\multicolumn{10}{l}{\textit{(c) Feature guidance}}\\
0.25 & $(0,0.3)$ & \cellcolor{guidanceorange}1 & VA & 100K & 32 & -- & -- & 37.27 & 40.39\\
0.25 & $(0,0.3)$ & \cellcolor{guidanceorange}5 & VA & 100K & 32 & -- & -- & 33.57 & 43.41\\
0.25 & $(0,0.3)$ & \cellcolor{guidanceorange}15 & VA & 100K & 32 & -- & -- & 27.77 & 52.08\\
0.25 & $(0,0.3)$ & \cellcolor{guidanceorange}30 & VA & 100K & 32 & -- & -- & 23.89 & \textbf{57.97}\\
0.25 & $(0,0.3)$ & \cellcolor{guidanceorange}45 & VA & 100K & 32 & -- & -- & 23.97 & 56.33\\
0.25 & $(0,0.3)$ & \cellcolor{guidanceorange}60 & VA & 100K & 32 & -- & -- & \textbf{23.83} & 56.89\\
\midrule
\multicolumn{10}{l}{\textit{(d) Teacher preparation}}\\
0.25 & $(0,0.3)$ & 30 & \cellcolor{teacherpink}Std. & 100K & 32 & -- & -- & 29.75 & 48.95\\
0.25 & $(0,0.3)$ & 30 & \cellcolor{teacherpink}VA & 100K & 32 & -- & -- & \textbf{23.89} & \textbf{57.97}\\
\midrule
\multicolumn{10}{l}{\textit{(e) Tuning duration}}\\
0.25 & $(0,0.3)$ & 30 & VA & \cellcolor{lengthpurple}25K & 32 & -- & -- & 29.11 & 50.97\\
0.25 & $(0,0.3)$ & 30 & VA & \cellcolor{lengthpurple}100K & 32 & -- & -- & 23.89 & 57.97\\
0.25 & $(0,0.3)$ & 30 & VA & \cellcolor{lengthpurple}200K & 32 & -- & -- & \textbf{23.15} & \textbf{58.50}\\
\midrule
\multicolumn{10}{l}{\textit{(f) Batch size}}\\
0.25 & $(0,0.3)$ & 30 & VA & 100K & \cellcolor{batchyellow}32 & 25.07 & 0.97 & 23.89 & \textbf{57.97}\\
0.25 & $(0,0.3)$ & 30 & VA & 100K & \cellcolor{batchyellow}256 & \textbf{25.20} & \textbf{0.81} & \textbf{23.82} & 57.82\\
\bottomrule
}
\setlength{\tabcolsep}{2pt}
\sbox{\guatablebox}{\begin{tabular}{rcrlrrrrrr}\guarows\end{tabular}}
\setlength{\tabcolsep}{\dimexpr 2pt+(\linewidth*100/\guashrink-\wd\guatablebox)/20\relax}
\resizebox{\linewidth}{!}{\begin{tabular}{rcrlrrrrrr}\guarows\end{tabular}}
\end{table}

Table~\ref{tab:ablations} compares tuning choices using SiT-B and a
common downstream evaluation.

\textbf{Alignment weight.}
Increasing the alignment weight initially improves generation,
but consistently degrades reconstruction, as shown by decreasing
PSNR and increasing rFID (Table~\ref{tab:ablations}(a)).
We choose $\lambda_V=0.25$ for SiT-B to balance generation and
reconstruction quality. For the main XL runs, we use $0.05$:
compared with $0.25$, this weight better preserves reconstruction
at the cost of a modest increase in downstream gFID
(Appendix~\ref{app:reconstruction}).

\textbf{Teacher timestep.}
Table~\ref{tab:ablations}(b) compares shared and independent timesteps.
Shared-timestep alignment yields gFID $35.02$. Sampling the teacher
timestep independently from the same full range,
$\tau\sim\mathcal U(0,1)$, improves gFID to $27.68$.
Restricting the independent teacher timestep to $(0,0.3)$ further
improves it to $23.89$, the best result among the tested intervals.
These comparisons support both independent timestep sampling and
the use of low-noise teacher targets.

\textbf{Feature guidance.}
Using conditional teacher features without guidance ($\omega=1$)
yields gFID $37.27$, providing no improvement over the original
VAE ($37.02$). Increasing $\omega$ to $30$ reduces gFID to $23.89$
(Table~\ref{tab:ablations}(c)), demonstrating the importance of
amplifying the class-conditional signal in the alignment targets.
Appendix~\ref{app:feature_guidance_cfg} reports results with CFG.

\textbf{Teacher preparation.}
Adding auxiliary velocity prediction during teacher preparation improves
downstream gFID from $29.75$ to $23.89$ compared with diffusion-only
preparation (Table~\ref{tab:ablations}(d)). We use this prepared teacher
in the remaining experiments.

\textbf{Tuning duration.}
Most gains occur between $25$K and $100$K iterations, with smaller gains
at $200$K (Table~\ref{tab:ablations}(e)); we therefore use $100$K.

\textbf{Batch size.}
Batch sizes $32$ and $256$ give similar gFID in the SiT-B configuration
(Table~\ref{tab:ablations}(f)), but batch size $256$ yields substantially
better reconstruction. We use batch size $32$ for the other ablations
to reduce computational cost.

\subsection{Latent Structure and Diffusion Features}
\label{sec:representations}
\begin{wraptable}{r}{0.49\linewidth}
\centering
\vspace{-2em}
\caption{\textbf{Representation analysis.} SRSS on VAE latents and LNC@10 (\%) on SiT-B features.}
\label{tab:representations}
\small
\begingroup\setlength{\tabcolsep}{3pt}
\def\guapanelrows{
\toprule VAE & SRSS$\uparrow$ & LNC@10$\uparrow$\\
\midrule
\multicolumn{3}{l}{\textit{$f8d4$}}\\
SD-VAE & 0.330 & 8.23\\
\rowcolor{oursorange}Ours & 0.512 & 13.42\\
\midrule
REPA-E\dinomark & 0.636 & 12.85\\
\midrule
\multicolumn{3}{l}{\textit{$f16d32$}}\\
IN-VAE & 0.320 & 10.35\\
\rowcolor{oursorange}Ours & 0.549 & 14.04\\
\midrule
REPA-E\dinomark & 0.542 & 14.18\\
\bottomrule}
\sbox{\guatablebox}{\begin{tabular}{lrr}\guapanelrows\end{tabular}}
\setlength{\tabcolsep}{\dimexpr3pt+(\linewidth-\wd\guatablebox)/6\relax}
\begin{tabular}{lrr}\guapanelrows\end{tabular}
\endgroup
\end{wraptable}

\begin{figure}[!t]
\centering
\includegraphics[width=\linewidth]{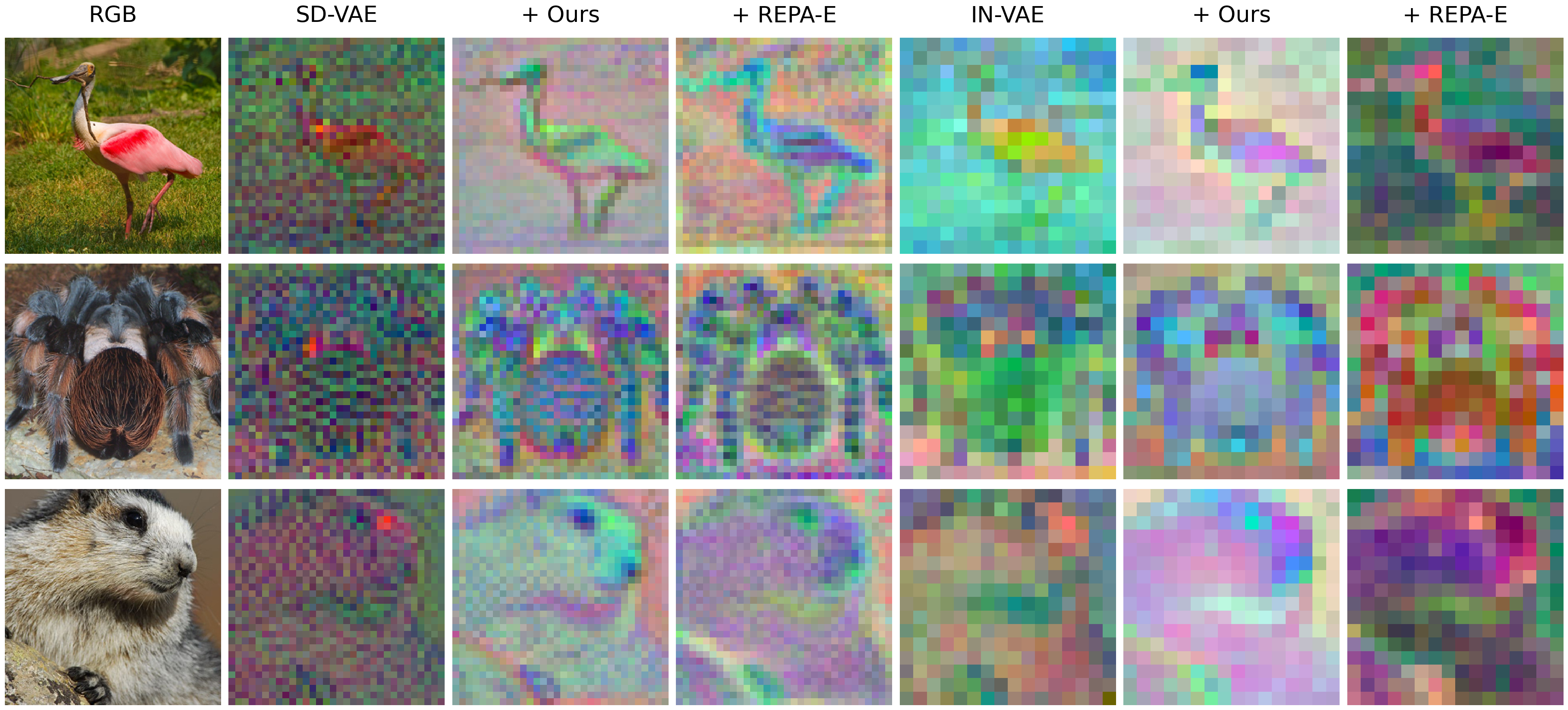}
\caption{\textbf{PCA comparison of VAE latents.} Three examples compare
the original SD-VAE and IN-VAE with our tuned VAEs and REPA-E. Both
\textit{+ Ours} columns use SiT-XL tuning with $\lambda_V=0.05$.}
\label{fig:latents_pca}
\end{figure}

We next examine how VAE tuning changes representations.
Figure~\ref{fig:latents_pca} shows more coherent object regions and less
checkerboard-like variation after tuning, particularly for SD-VAE.

Semantic-region self-similarity (SRSS) measures similarity within object
regions relative to their surroundings \citep{singh2025irepa}.
Prior work links this spatial structure to downstream generation quality
\citep{zhong2026diffusability}. Table~\ref{tab:representations} shows that
SRSS increases from $0.330$ to $0.512$ for SD-VAE and from $0.320$ to
$0.549$ for IN-VAE, supporting the visual evidence of stronger spatial
structure. Without DINO supervision, our tuning substantially narrows the SRSS gap to REPA-E on $f8d4$ and closes it on $f16d32$.

\begin{table}[!t]
\centering
\normalcolor
\begin{minipage}[t]{0.49\linewidth}
\centering
\small
\caption{\textbf{Text-to-image generation on COCO.}
gFID of MMDiT trained on frozen VAEs.\strut}
\label{tab:text_to_image}
\setlength{\tabcolsep}{3pt}
\def\guapanelrows{
\toprule
Frozen VAE & gFID w/o CFG $\downarrow$ & gFID w/ CFG $\downarrow$ \\
\midrule
SD-VAE & 12.82 & 5.92 \\
REPA-E\dinomark & 6.62 & 4.96 \\
\rowcolor{oursorange}Ours & 6.76 & 4.95 \\
\bottomrule
}
\sbox{\guatablebox}{\begin{tabular}{lrr}\guapanelrows\end{tabular}}
\setlength{\tabcolsep}{\dimexpr3pt+(\linewidth-\wd\guatablebox)/6\relax}
\begin{tabular}{lrr}\guapanelrows\end{tabular}
\end{minipage}\hfill
\begin{minipage}[t]{0.49\linewidth}
\centering
\small
\caption{\textbf{Student training.} Total counts updates across all stages.\strut}
\label{tab:training_strategies}
\small\setlength{\tabcolsep}{3pt}
\begingroup
\setlength{\tabcolsep}{3pt}
\def\guapanelrows{
\toprule Student & Teacher VAE & Total & gFID$\downarrow$\\
\midrule
New & Tuned & 900K & 17.17\\
New & SD-VAE & 700K & 17.83\\
\rowcolor{oursorange}Continued & -- & 600K & 17.47\\
\bottomrule
}
\sbox{\guatablebox}{\begin{tabular}{llrr}\guapanelrows\end{tabular}}
\setlength{\tabcolsep}{\dimexpr3pt+(\linewidth-\wd\guatablebox)/8\relax}
\begin{tabular}{llrr}\guapanelrows\end{tabular}
\endgroup

\end{minipage}
\end{table}

\begin{table}[!t]
\centering
\caption{\textbf{Short-run ImageNet-$256$ generation.} All generators are trained
for $80$ epochs ($400$K updates) and evaluated without guidance.
--: not reported; sources in Appendix~\ref{app:evaluation}.}
\label{tab:short_generation}
\small
\begin{minipage}[t]{0.485\linewidth}
\centering\textbf{(a) SiT-B, $f8d4$}\par\smallskip
\begingroup
\setlength{\tabcolsep}{3pt}
\def\guapanelrows{
\toprule Method & gFID$\downarrow$ & IS$\uparrow$\\\midrule
SiT & 37.02 & 40.38\\
\mbox{Self-Transcendence} & 20.49 & 70.57\\
\rowcolor{oursorange}Ours & \textbf{17.47} & \textbf{75.48}\\\bottomrule
}
\sbox{\guatablebox}{\begin{tabular}{lrr}\guapanelrows\end{tabular}}
\setlength{\tabcolsep}{\dimexpr3pt+(\linewidth-\wd\guatablebox)/6\relax}
\begin{tabular}{lrr}\guapanelrows\end{tabular}
\endgroup
\end{minipage}\hfill
\begin{minipage}[t]{0.485\linewidth}
\centering\textbf{(b) SiT-XL, $f8d4$}\par\smallskip
\begingroup
\setlength{\tabcolsep}{3pt}
\def\guapanelrows{
\toprule Method & gFID$\downarrow$ & IS$\uparrow$\\\midrule
SiT & 17.46 & 75.53\\
\mbox{Self-Transcendence} & 7.51 & --\\
\rowcolor{oursorange}Ours & \textbf{6.79} & \textbf{131.14}\\\bottomrule
}
\sbox{\guatablebox}{\begin{tabular}{lrr}\guapanelrows\end{tabular}}
\setlength{\tabcolsep}{\dimexpr3pt+(\linewidth-\wd\guatablebox)/6\relax}
\begin{tabular}{lrr}\guapanelrows\end{tabular}
\endgroup
\end{minipage}
\end{table}

We also measure latent neighbor consistency (LNC)
\citep{zhong2026diffusability} on SiT-B features rather than VAE latents.
LNC@10 is the mean fraction of same-class neighbors among the ten
nearest feature vectors. With a fixed layer and noise level, it rises
from $8.23\%$ to $13.42\%$ for SD-VAE and from $10.35\%$ to $14.04\%$
for IN-VAE, reaching values close to REPA-E. Thus the spatial changes
in VAE latents are accompanied by more class-consistent neighborhoods
in the downstream generator. Appendix~\ref{app:representations}
specifies both protocols.
\finishwrap

\begingroup
\subsection{Transfer to Text-to-Image Generation}
\label{sec:t2i}

We next evaluate whether the improvements from ImageNet tuning
transfer to text-to-image generation on COCO. We train an MMDiT
from scratch on each frozen VAE using only the diffusion loss,
without further VAE tuning. We compare our tuned SD-VAE from Table~\ref{tab:vae_quality}(a)
with SD-VAE and REPA-E, with and without CFG. Training and evaluation details are provided
in Appendix~\ref{app:t2i}.

Table~\ref{tab:text_to_image} shows that our tuned VAE improves over SD-VAE and achieves performance close to REPA-E
both with and without CFG. These results show that the benefits of
VAE tuning transfer to a different dataset, conditioning modality,
and generator architecture.
\par
\endgroup

\subsection{From VAE Tuning to a Generative Model}
\label{sec:synergy}
Beyond providing a reusable VAE, our tuning procedure also produces
a diffusion student that can be continued for generation.
Table~\ref{tab:training_strategies} compares continuing this student
with training a new Self-Transcendence student on the tuned VAE.
Continuation gives nearly the best gFID with fewer training stages,
so we freeze the VAE and continue the existing student using diffusion
loss alone. Totals include teacher preparation and E2E;
Appendix~\ref{app:training_details} gives the schedules.
\par
\finishwrap

\textbf{Generation at different training durations.} Table~\ref{tab:short_generation} compares generation after $80$ epochs.
Our procedure improves on SiT and Self-Transcendence with both B and
XL backbones.

\begin{table}[!tbp]
\centering
\caption{\textbf{Class-conditional ImageNet-$256$ generation.}
Epochs count generator training; $80$ epochs corresponds to $400$K updates.
$\dagger$: our evaluation of released weights;
$\mathrm{AG}$: AutoGuidance. Other guided entries use CFG.
--: not reported. Sources and sampling protocols are in Appendix~\ref{app:evaluation}.}
\label{tab:final_generation}
\small\setlength{\tabcolsep}{1.5pt}
\def\guafinalrows{\toprule
& & & \multicolumn{2}{c}{Without guidance} & \multicolumn{5}{c}{With guidance}\\
\cmidrule(lr){4-5}\cmidrule(l){6-10}
Method & Latents & Epochs & gFID$\downarrow$ & IS$\uparrow$ & gFID$\downarrow$ & sFID$\downarrow$ & IS$\uparrow$ & Prec.$\uparrow$ & Rec.$\uparrow$\\
\midrule
\multicolumn{10}{l}{\textit{Without DINO supervision}}\\
DiT-XL/2 & $f8d4$ & 1400 & 9.62 & 121.50 & 2.27 & 4.60 & 278.24 & 0.83 & 0.57\\
SiT-XL/2 & $f8d4$ & 1400 & 8.61 & 131.70 & 2.06 & 4.50 & 270.30 & 0.82 & 0.59\\
MaskDiT & $f8d4$ & 1600 & 5.69 & 177.99 & 2.28 & 5.67 & 276.56 & 0.80 & 0.61\\
U-ViT-H/2 & $f8d4$ & 400 & 6.58 & -- & 2.29 & 5.68 & 263.88 & 0.82 & 0.57\\
SRA & $f8d4$ & 400 & -- & -- & 1.85 & 4.50 & 297.20 & 0.82 & 0.61\\
SRA2 & $f8d4$ & 400 & 7.00 & -- & 1.84 & 4.50 & 306.50 & 0.82 & 0.61\\
LayerSync & $f8d4$ & 800 & 6.32 & -- & 1.49 & -- & 285.20 & 0.80 & 0.63\\
\mbox{Self-Transcendence} & $f8d4$ & 400 & $6.73^{\dagger}$ & $149.27^{\dagger}$ & 1.44 & 4.85 & 311.30 & 0.79 & 0.66\\
\rowcolor{oursorange}Ours & $f8d4$ & 400 & 4.97 & 160.41 & 1.62 & 4.96 & 293.73 & 0.795 & 0.638\\
\rowcolor{oursorange}Ours & $f8d4$ & 800 & 3.76 & 180.86 & 1.58 & 4.80 & 284.06 & 0.793 & 0.640\\
\rowcolor{oursorange}Ours & $f16d32$ & 160 & 4.36 & 155.56 & 1.62 & 4.77 & 259.99 & 0.788 & 0.641\\
\rowcolor{oursorange}Ours & $f16d32$ & 800 & \textbf{3.73} & 176.47 & \textbf{1.43} & 4.90 & 289.90 & 0.783 & 0.646\\
\midrule
\multicolumn{10}{l}{\textit{With DINO supervision}}\\
REPA\dinomark & $f8d4$ & 800 & 5.90 & 157.80 & 1.42 & 4.70 & 305.70 & 0.80 & 0.65\\
REPA-E\dinomark & $f16d32$ & 800 & 1.83 & 217.30 & \textbf{1.26} & -- & 314.90 & -- & --\\
VA-VAE\dinomark & $f16d32$ & 800 & 2.17 & 205.60 & 1.35 & 4.15 & 295.30 & 0.79 & 0.65\\
RAE$^{*\mathrm{AG}}$ & $16^2\!\times\!768$ & 800 & \textbf{1.60} & 242.70 & 1.28 & -- & 262.90 & -- & --\\
\bottomrule}
\sbox{\guatablebox}{\begin{tabular}{lcrrrrrrrr}\guafinalrows\end{tabular}}
\setlength{\tabcolsep}{\dimexpr1.5pt+(\textwidth-\wd\guatablebox)/20\relax}
\begin{tabular}{lcrrrrrrrr}\guafinalrows\end{tabular}
\end{table}

Table~\ref{tab:final_generation} compares complete generative systems
at longer training durations. At $400$ epochs, our SD-VAE system
reaches gFID $4.97$ without sampling guidance, compared with $6.73$
for Self-Transcendence. It also improves on SRA2 at $400$ epochs and
LayerSync at $800$ epochs, as well as DiT, SiT, and MaskDiT trained
for substantially longer. Our IN-VAE system reaches unguided gFID
$4.36$ at $160$ epochs.
Extending training to $800$ epochs improves unguided gFID to
$3.76$ with SD-VAE and $3.73$ with IN-VAE.

With CFG, the $800$-epoch runs achieve gFID $1.58$ and $1.43$,
respectively. At the matched $400$-epoch budget with SD-VAE,
Self-Transcendence remains stronger ($1.44$ versus $1.62$).
The $800$-epoch IN-VAE result is numerically lower, but uses a different
latent format and a longer training schedule. DINO-supervised systems retain
an advantage: REPA improves diffusion training, while REPA-E, VA-VAE,
and RAE also change the representation used for generation. We use
random-label sampling results where available and explicitly mark
RAE's AutoGuidance. These are
system-level comparisons, complementing the controlled frozen-VAE
experiments in Table~\ref{tab:vae_quality}.

\FloatBarrier
\section{Conclusion}
We introduced \gvae{}, an end-to-end VAE tuning method that uses
CFG-enhanced diffusion features as supervision without an external
visual encoder. Independently sampling low-noise teacher timesteps
improves the effectiveness of this supervision. Across VAEs with
different latent shapes, the tuned latent spaces improve generation
with diffusion models trained from scratch. Increasing the teacher
and student sizes further improves the resulting VAEs, suggesting
that stronger diffusion models can provide more effective supervision
for tokenizer learning.

These results make diffusion representations a practical alternative
to external visual features for improving VAE latent spaces.
The tuning procedure produces both a reusable VAE and a diffusion
model that can be continued with the diffusion objective alone.
It therefore supports subsequent training of new generators while
also providing a direct path to a competitive generative model
through continuation of the existing model.

\FloatBarrier
\section*{AI Use Statement}
We used generative AI tools to assist with polishing and revising
the manuscript, including improving grammar, clarity, readability,
and the organization of the text. All AI-assisted text was reviewed
and edited by the authors, who take full responsibility for the
final content of the paper.

\FloatBarrier
\bibliographystyle{iclr2027_conference}
\bibliography{references}

\newpage
\FloatBarrier
\appendix
\section{Training Details}
\label{app:training_details}
\paragraph{Teacher preparation.}
We reproduce the first-stage training objective from the released
Self-Transcendence implementation \citep{sun2026selftranscendence}.
For the linear interpolant, both the final prediction and an auxiliary
head at an early diffusion layer predict the velocity $\epsilon-z$:
\begin{equation}
 \mathcal L_{\mathrm{aux}}
 =\mathbb E\!\left[\frac{1}{N}
 \left\|q(h_{\ell}(z_t,t,y))-(\epsilon-z)\right\|_2^2\right],
 \qquad
 \mathcal L_{\mathrm{teacher}}=\Ldiff+0.5\mathcal L_{\mathrm{aux}}.
\end{equation}
Here $q$ maps the intermediate features to a latent-shaped prediction,
and $N$ is the number of scalar latent entries. We attach the auxiliary head at $\ell=\operatorname{round}(L/6)$
for a backbone of depth $L$. This gives layer $2$ for SiT-B,
layer $4$ for SiT-L, and layer $5$ for both SiT-XL and SiT-2XL. All layer
indices count from one. Teacher preparation lasts $200$K iterations;
the teacher and its VAE are then frozen throughout VAE tuning.

\paragraph{Backbones and alignment layers.}
SiT-B has depth $12$, hidden size $768$, and $12$ heads.
SiT-L has depth $24$, hidden size $1024$, and $16$ heads.
SiT-XL has depth $28$, hidden size $1152$, and $16$ heads. SiT-2XL has depth $32$,
hidden size $1536$, patch size $2$, and $16$ heads, with
$1{,}379{,}641{,}872$ SiT parameters. During VAE tuning, teacher and
student use the same backbone size. We extract student features at $\operatorname{round}(L/2)$ and
teacher features at $\operatorname{round}(2L/3)$, giving layer pairs
$(6,8)$, $(12,16)$, $(14,19)$, and $(16,21)$ for B, L, XL, and 2XL,
respectively.
Feature guidance uses $\omega=30$ and the student alignment weight is
$\lambda_S=0.5$, following the SiT settings of Self-Transcendence
\citep[Table~5]{sun2026selftranscendence}. The feature alignment loss
is MSE on the projected student features and teacher targets, without
unit-norm feature normalization.

\paragraph{VAE tuning configurations.}
Table~\ref{tab:training_configs} lists the tuning settings. Unless varied
in an ablation, tuning lasts $100$K iterations and teacher timesteps
are sampled from $(0,0.3)$. The main XL configurations use
$\lambda_V=0.05$ to preserve reconstruction quality; the B-model
configuration uses $0.25$. The $f8d4$ scaling comparison instead fixes
$\lambda_V=0.25$ and batch size $256$ across all backbone sizes.
SD-VAE supplies the teacher latents for both selected VAE formats.
The final IN-VAE configuration uses SiT-XL, batch size $256$, and
$\lambda_V=0.05$; its VAE is shared by the frozen-VAE, representation,
and $160$-epoch generative configurations.
\begin{table}[!htbp]
\centering
\caption{\textbf{Training configurations for VAE tuning.} Student and Teacher denote VAEs; SiT gives both diffusion model sizes.}
\label{tab:training_configs}

\small\setlength{\tabcolsep}{5pt}
\begin{tabular*}{\linewidth}{@{\extracolsep{\fill}}llcrr@{}}
\toprule
Student VAE & Teacher VAE & SiT & BS & $\lambda_V$\\
\midrule
SD-VAE & SD-VAE & B / L / XL / 2XL & 256 & 0.25\\
SD-VAE & SD-VAE & XL & 256 & 0.05\\
SD-VAE & SD-VAE & B & 32 & 0.25\\
IN-VAE & IN-VAE & B & 32 & 0.25\\
IN-VAE & SD-VAE & B & 32 & 0.25\\
IN-VAE & SD-VAE & XL & 256 & 0.25\\
IN-VAE & SD-VAE & XL & 256 & 0.05\\
\bottomrule
\end{tabular*}
\end{table}

\paragraph{Latent shapes and noise.}
At $256\times256$ resolution, SD-VAE produces $32\times32\times4$
latents and IN-VAE produces $16\times16\times32$ latents. Patch sizes
$2$ and $1$ give matching $16\times16$ diffusion-token grids.
Same-shape SD-VAE tuning shares the Gaussian draw between teacher
and student. Cross-VAE tuning uses independent Gaussian draws for the
two latent shapes. Conditional and unconditional teacher passes use
the same noisy teacher input. Teacher and student timesteps are
independent except in the shared-timestep ablation.

\paragraph{Optimization and alternating updates.}
Each iteration first updates the VAE, then the discriminator, and then
the student. The VAE update holds student parameters fixed while
differentiating through the student's activations. The student update
uses detached VAE latents. Both updates use the same minibatch,
sampled timesteps, Gaussian draws, and teacher target. Student latent
batch-normalization statistics are fixed during the VAE update and
updated during the student update; teacher statistics remain fixed.
The VAE encoder and decoder retain the pixel, perceptual, adversarial,
and KL objectives of the REPA-E tuning recipe \citep{leng2025repae}.
The discriminator is optimized separately. We use AdamW with learning
rate $10^{-4}$, $(\beta_1,\beta_2)=(0.9,0.999)$, no weight decay, and
gradient clipping at norm $1$. The student uses class-label dropout
with probability $0.1$ during training; the teacher's conditional and
unconditional passes use explicit class and null labels.

\paragraph{Choosing the teacher VAE for IN-VAE tuning.}
\label{app:teacher_vae}
With SiT-B teacher and student, batch size $32$, and $\lambda_V=0.25$,
an SD-VAE teacher yields downstream gFID $21.06$, compared with
$26.12$ for an IN-VAE teacher. This controlled comparison motivates
using SD-VAE for the teacher. Table~\ref{tab:invae_configs} gives the
corresponding runs and the XL configurations. B and XL differ in
batch size, so these results are not a controlled scaling comparison.
With the XL backbone and SD-VAE teacher, reducing $\lambda_V$ from
$0.25$ to the selected $0.05$ changes downstream gFID from $18.88$
to $19.30$.
\begin{table}[!htbp]
\centering\small
\caption{\textbf{IN-VAE tuning configurations.} SiT gives both teacher and
student sizes during tuning. Evaluation uses a new SiT-B/1 trained on
each frozen VAE for $400$K iterations.}
\label{tab:invae_configs}
\setlength{\tabcolsep}{7pt}
\begin{tabular*}{\linewidth}{@{\extracolsep{\fill}}llrrrr@{}}
\toprule
Teacher VAE & SiT & BS & $\lambda_V$ & gFID$\downarrow$ & IS$\uparrow$\\
\midrule
IN-VAE & B & 32 & 0.25 & 26.12 & 53.26\\
SD-VAE & B & 32 & 0.25 & 21.06 & 63.24\\
\midrule
SD-VAE & XL & 256 & 0.25 & 18.88 & 67.98\\
SD-VAE & XL & 256 & 0.05 & 19.30 & 67.09\\
\bottomrule
\end{tabular*}
\end{table}

\paragraph{Ablation settings.}
Both teacher variants in Table~\ref{tab:ablations} are SiT-B models
trained for $200$K iterations. VA denotes auxiliary velocity supervision
and Std. denotes the final diffusion objective alone. Teacher intervals
$(a,b)$ denote independent $\tau\sim\mathcal U(a,b)$; $\tau=t$ shares
the student's timestep. The alignment-weight sweep uses $(0,0.1)$;
the duration and batch-size sweeps use $(0,0.3)$. The timestep sweep compares shared timesteps with independent
sampling over the full range $(0,1)$ and the listed restricted intervals.

\paragraph{From the tuned VAE to a generator.}
All strategies in Table~\ref{tab:training_strategies} start with $200$K
teacher-preparation updates and $100$K E2E updates, using the same
SiT-B-tuned VAE with tuning batch size $256$. Training a new teacher
on this VAE adds $200$K updates, followed by $400$K for a new student:
$900$K updates in total. Reusing the original teacher skips the extra
teacher training, giving $700$K in total. Both new students use feature
alignment for their first $100$K updates with shared teacher/student
timesteps, followed by $300$K diffusion-only updates; the VAE is frozen.
Continuation instead adds $300$K diffusion-only updates to the E2E
student, giving $600$K in total. These totals count optimization steps
across stages and do not imply equal FLOPs or wall-clock cost.

\begingroup
\makeatletter
\long\def\@makecaption#1#2{%
  \vskip\abovecaptionskip
  \noindent #1: #2\par
  \vskip\belowcaptionskip}
\makeatother
\paragraph{Training cost.}
\label{app:training_cost}
We measure all training stages using SiT-XL/2 and SD-VAE on two
NVIDIA A100 80\,GB GPUs, with BF16 precision and an effective batch
size of $256$. Both E2E methods use $16$ gradient-accumulation steps;
each reported iteration includes all $16$ microsteps. Timings cover
training computation only, excluding data reading and transfer to
the GPU, latent-cache preparation, and checkpoint saving.
Table~\ref{tab:training_throughput} reports the throughput of each stage.

\begin{table}[htbp]
\centering
\small
\caption{\textbf{Training throughput.} All stages use the same
hardware and effective batch size.}
\label{tab:training_throughput}
\setlength{\tabcolsep}{5pt}
\renewcommand{\arraystretch}{1.12}
\begin{tabular*}{\linewidth}{@{\extracolsep{\fill}}llr@{}}
\toprule
Method & Training stage & Iterations/second $\uparrow$ \\
\midrule
Ours / Self-Transcendence & Teacher preparation & 1.185 \\
Ours & E2E VAE tuning & 0.136 \\
REPA-E & E2E VAE tuning & 0.145 \\
Self-Transcendence & Student with feature alignment & 0.807 \\
All three methods & Student, diffusion loss only & 1.20 \\
\bottomrule
\end{tabular*}
\end{table}

\paragraph{E2E iteration breakdown.}
VAE training, reconstruction losses, and the discriminator account
for approximately $70\%$ of the E2E iteration in both methods
(Table~\ref{tab:e2e_breakdown}).
The additional diffusion-teacher computation makes our E2E
iteration approximately $7\%$ longer than a REPA-E iteration.

\begin{table}[htbp]
\centering
\small
\caption{\textbf{E2E step time.} VAE update includes reconstruction losses and the discriminator.}
\label{tab:e2e_breakdown}
\renewcommand{\arraystretch}{1.12}
\begin{tabular*}{\linewidth}{@{\extracolsep{\fill}}lrr@{}}
\toprule
E2E component & Ours (s) & REPA-E (s) \\
\midrule
VAE update & 4.97 & 4.96 \\
SiT update (with alignment) & 1.83 & 1.81 \\
Teacher / DINO forward & 0.57 & 0.11 \\
\midrule
\textbf{Total} & \textbf{7.36} & \textbf{6.89} \\
\bottomrule
\end{tabular*}
\end{table}

\paragraph{VAE preparation cost.}
Table~\ref{tab:tuning_time} reports the estimated compute required to obtain
each tuned VAE, using the measured stage throughputs. Our method uses $200$K teacher-preparation
updates followed by $100$K E2E updates, while REPA-E uses $400$K
E2E updates.

\begin{table}[htbp]
\centering
\small
\caption{\textbf{VAE preparation cost.} Updates / GPU-hours,
including teacher preparation.}
\label{tab:tuning_time}
\setlength{\tabcolsep}{7pt}
\renewcommand{\arraystretch}{1.12}
\begin{tabular*}{\linewidth}{@{\extracolsep{\fill}}lrrr@{}}
\toprule
Method & Teacher preparation & E2E tuning & Total GPU-hours $\downarrow$ \\
\midrule
REPA-E & --- & 400K / 1,530.6 & \textbf{1,530.6} \\
Ours & 200K / 93.8 & 100K / 408.8 & \textbf{502.6} \\
\bottomrule
\end{tabular*}
\end{table}

Under these schedules, our method provides an estimated
$3.05\times$ speedup over REPA-E in VAE preparation,
including teacher training. This comparison covers the cost of
obtaining the tuned VAE, before downstream generator training.

\paragraph{Total training budget.}
Table~\ref{tab:total_training_cost} estimates the total compute when VAE
preparation is followed by generator training, with each final
student receiving $2$M updates. Our generative system combines
the tuned VAE with continuation of the jointly trained SiT
student.
All estimates use the same SiT-XL/2 and SD-VAE configuration.

\begin{table}[htbp]
\centering
\small
\caption{\textbf{Total training cost at 2M student updates.} Entries show
updates / GPU-hours, estimated from measured step times.}
\label{tab:total_training_cost}
\setlength{\tabcolsep}{4pt}
\renewcommand{\arraystretch}{1.12}
\begin{tabular*}{\linewidth}{@{\extracolsep{\fill}}lrrrr@{}}
\toprule
Generative system & \shortstack{Teacher\\preparation}
& \shortstack{VAE\\tuning}
& \shortstack{Subsequent\\student training}
& \shortstack{Total\\GPU-hours $\downarrow$} \\
\midrule
REPA-E + SiT & --- & 400K / 1,530.6 & 2M / 924.5 & \textbf{2,455.1} \\
Ours & 200K / 93.8 & 100K / 408.8 & 1.9M / 877.8 & \textbf{1,380.4} \\
Self-Transcendence & 200K / 93.8 & --- & 2M / 937.0 & \textbf{1,030.7} \\
\bottomrule
\end{tabular*}
\end{table}

Our system's $100$K E2E updates contribute to the final student's
training, followed by $1.9$M diffusion-only updates. Following
\citet{sun2026selftranscendence}, Self-Transcendence trains its SiT-XL student
for $50$K updates with feature alignment and $1.95$M with the diffusion
loss alone. Totals include diffusion-teacher preparation and exclude
the original pretraining of SD-VAE and DINO.

Compared with Self-Transcendence, our generative system requires
approximately $34\%$ more compute at $2$M student updates; this
overhead decreases to $18\%$ at $4$M updates.
\par
\endgroup
\FloatBarrier

\section{Evaluation Protocols and Baseline Sources}
\label{app:evaluation}
\paragraph{Frozen-VAE evaluation.}
\leavevmode For ImageNet, we train a new diffusion model for $400$K iterations on each frozen
VAE, using only the diffusion objective. SiT-B/2 and SiT-XL/2 evaluate f8d4 latents, while SiT-B/1 and SiT-XL/1 evaluate f16d32 latents, giving the same
$16\times16$ token grid. These runs assess the tuned VAE independently
of the E2E student. The EQ-VAE gFID is from
\citet[Table~2]{kouzelis2025eqvae}; the remaining frozen-VAE results are
our evaluations. Table~\ref{tab:vae_quality_is} provides IS for the same
models, supplementing the gFID values in Table~\ref{tab:vae_quality}.
\begin{table}[!htbp]
\centering
\caption{\textbf{Inception scores for frozen-VAE comparisons.} These scores
supplement Table~\ref{tab:vae_quality} for the same selected VAEs and generators.}
\label{tab:vae_quality_is}
\small
\begin{minipage}[t]{0.315\linewidth}
\centering\textbf{(a) SiT-B, $f8d4$}\par\smallskip
\begingroup
\setlength{\tabcolsep}{3pt}
\def\guapanelrows{
\toprule VAE & IS$\uparrow$\\\midrule
SD-VAE & 40.38\\
EQ-VAE & --\\
\rowcolor{oursorange}Ours & 61.37\\
\midrule
REPA-E\dinomark & 65.37\\\bottomrule
}
\sbox{\guatablebox}{\begin{tabular}{lr}\guapanelrows\end{tabular}}
\setlength{\tabcolsep}{\dimexpr3pt+(\linewidth-\wd\guatablebox)/4\relax}
\begin{tabular}{lr}\guapanelrows\end{tabular}
\endgroup
\end{minipage}\hfill
\begin{minipage}[t]{0.315\linewidth}
\centering\textbf{(b) SiT-B, $f16d32$}\par\smallskip
\begingroup
\setlength{\tabcolsep}{3pt}
\def\guapanelrows{
\toprule VAE & IS$\uparrow$\\\midrule
IN-VAE & 36.95\\
\rowcolor{oursorange}Ours & 67.09\\
\midrule
REPA-E\dinomark & 72.21\\
VA-VAE\dinomark & 81.39\\\bottomrule
}
\sbox{\guatablebox}{\begin{tabular}{lr}\guapanelrows\end{tabular}}
\setlength{\tabcolsep}{\dimexpr3pt+(\linewidth-\wd\guatablebox)/4\relax}
\begin{tabular}{lr}\guapanelrows\end{tabular}
\endgroup
\end{minipage}\hfill
\begin{minipage}[t]{0.315\linewidth}
\centering\textbf{(c) SiT-XL}\par\smallskip
\begingroup
\setlength{\tabcolsep}{3pt}
\def\guapanelrows{
\toprule Latent & VAE & IS$\uparrow$\\\midrule
\multirow{2}{*}{$f8d4$} & SD-VAE & 75.53\\
& \cellcolor{oursorange}Ours & \cellcolor{oursorange}\textbf{103.10}\\
\midrule
\multirow{2}{*}{$f16d32$} & IN-VAE & 71.08\\
& \cellcolor{oursorange}Ours & \cellcolor{oursorange}\textbf{111.16}\\\bottomrule
}
\sbox{\guatablebox}{\begin{tabular}{clr}\guapanelrows\end{tabular}}
\setlength{\tabcolsep}{\dimexpr3pt+(\linewidth-\wd\guatablebox)/6\relax}
\begin{tabular}{clr}\guapanelrows\end{tabular}
\endgroup
\end{minipage}
\end{table}

\begingroup

\paragraph{Generation with CFG.}
\phantomsection
\label{app:frozen_vae_cfg}

We evaluate the same SiT-B/2 models from
Table~\ref{tab:vae_quality}(a), trained from scratch on frozen
VAEs using only the diffusion loss.
CFG settings are selected separately for each model using
the same search budget and parameter ranges.

\begin{table}[htbp]
\centering
\small
\caption{\textbf{Generation with CFG using frozen VAEs.}
Results for the SiT-B/2 models from
Table~\ref{tab:vae_quality}(a), evaluated on $50$K samples.}
\label{tab:frozen_vae_cfg}
\begin{tabular}{lrrrr}
\toprule
Frozen VAE
& \shortstack{gFID without\\CFG $\downarrow$}
& \shortstack{gFID with\\CFG $\downarrow$}
& \shortstack{CFG\\scale}
& \shortstack{Guidance\\interval} \\
\midrule
SD-VAE          & 37.02 & 6.30 & 3.0 & $[0,0.72]$ \\
REPA-E\dinomark & 20.50 & 3.69 & 2.9 & $[0,0.73]$ \\
\rowcolor{oursorange}Ours & 22.53 & 4.01 & 3.0 & $[0,0.74]$ \\
\bottomrule
\end{tabular}
\end{table}

With CFG, our tuned VAE improves over SD-VAE and achieves performance
close to REPA-E.

\paragraph{Feature-guidance ablation.}
\phantomsection
\label{app:feature_guidance_cfg}
We extend the ablation in Table~\ref{tab:ablations}(c) by evaluating the
same frozen VAEs with CFG (Table~\ref{tab:feature_guidance_cfg}).
Increasing $\omega$ from $1$ to $30$ improves gFID both without CFG
($37.27$ to $23.89$) and
with CFG ($6.47$ to $4.35$). Further increasing $\omega$ provides no
consistent improvement. Across $\omega=30,45,60$, gFID ranges from
$23.83$ to $23.97$ without CFG and from $4.35$ to $4.63$ with CFG.
These results indicate that sufficiently strong feature guidance benefits
both sampling settings, while performance is relatively insensitive to
its precise weight within the tested range from $30$ to $60$. This trend
is consistent with Self-Transcendence \citep[Table~5]{sun2026selftranscendence},
where $\omega=15$ also underperforms $\omega=30$, while FID varies by
only $0.66$ across $\omega=30,45,60$.

\begin{table}[htbp]
\centering
\small
\caption{\textbf{Feature guidance with and without sampling CFG.}
Each entry is gFID on $50$K samples. Guided sampling uses scale $3.0$
over $[0,0.74]$.}
\label{tab:feature_guidance_cfg}
\begin{tabular*}{\linewidth}{@{\extracolsep{\fill}}lrrrrrr@{}}
\toprule
Feature-guidance weight $\omega$ & 1 & 5 & 15 & 30 & 45 & 60\\
\midrule
gFID without CFG $\downarrow$ & 37.27 & 33.57 & 27.77 & 23.89 & 23.97 & \textbf{23.83}\\
gFID with CFG $\downarrow$ & 6.47 & 5.59 & 4.54 & \textbf{4.35} & 4.63 & 4.39\\
\bottomrule
\end{tabular*}
\end{table}

\par
\endgroup

\paragraph{Generation and training budgets.}
Our ImageNet generation metrics use $50{,}000$ images with uniformly random
class labels, the ADM evaluator \citep{dhariwal2021diffusion}, and
$250$-step Euler--Maruyama sampling. Our $400$-epoch SD-VAE and $160$-epoch IN-VAE results
in Table~\ref{tab:final_generation}
use CFG scale $1.6$ over timestep interval $(0,0.825)$.
For the $800$-epoch runs, we use CFG scale $1.5$ over
$(0,0.825)$ for SD-VAE and scale $1.65$ over $(0,0.855)$ for IN-VAE.
Feature CFG is used
during training even when sampling is unguided. In the $80$-epoch
rows, the student receives $100$K E2E updates and $300$K diffusion-only
updates. Reported epochs count generator/student training and exclude
teacher preparation and autoencoder training; they are not a measure
of matched compute across different methods. Results are single
evaluations rather than averages over training seeds.

\paragraph{Label sampling and guidance.}
Random-label sampling draws each of the $50{,}000$ class labels
independently. Class-balanced sampling instead generates exactly
$50$ images per class. The latter can yield lower gFID, as documented
by REPA-E and RAE \citep{leng2025repae,zheng2025rae}. We therefore use
random-label results for REPA-E and RAE from the Random columns of
\citet[Table~14]{zheng2025rae}, rather than their class-balanced headline
results. That table reports gFID and IS; we leave the other metrics
unreported instead of mixing sampling protocols within a row.
For VA-VAE we use the original paper's results; its official sampling
procedure draws class labels independently, unlike the class-balanced
VA-VAE re-evaluation in REPA-E. The remaining methods retain their
published sampling and guidance settings; this is not a standardized
re-evaluation of every system. RAE's guided result uses AutoGuidance,
marked $\mathrm{AG}$; other guided rows use CFG, including guidance
schedules where used in the source paper.

\paragraph{Diffusion baselines.}
DiT and SiT use the XL/2 architecture and their published $1400$-epoch
results \citep{peebles2023dit,ma2024sit}. For SiT we use the random-label
gFID and IS values also recorded by \citet[Table~14]{zheng2025rae}.
MaskDiT uses the $256\times256$ results in
\citet[Table~1]{zheng2024maskdit}; its $1600$-epoch budget follows the
reporting convention in \citet[Table~4]{yu2025repa}, and the final
checkpoints include the unmasking fine-tuning described by MaskDiT.
U-ViT uses the H/2 model \citep{bao2023uvit}, trained for $500$K
updates at batch size $1024$ (approximately $400$ epochs). These methods provide
baselines without external DINO supervision.

\paragraph{Self-contained alignment baselines.}
SRA uses its $400$-epoch guided result \citep[Table~3]{jiang2025sra}. SRA2 uses its $2$M-step unguided and
$400$-epoch guided results \citep[Tables~2--3]{wang2026sra2}.
LayerSync uses the $800$-epoch SDE result without CFG and its
scheduled-CFG result \citep[Tables~1--2]{haghighi2025layersync}.
Self-Transcendence's $80$-epoch results and $400$-epoch guided metrics
are taken from \citet{sun2026selftranscendence}. For its $400$-epoch
unguided result, we sample the authors' released $2$M-step checkpoint
without retraining; these entries are marked $\dagger$.

\paragraph{DINO-supervised systems.}
REPA uses SiT-XL/2 and its $800$-epoch results, with the guidance
interval setting for guided generation \citep{yu2025repa}.
REPA-E uses its tuned $f16d32$ VAE and SiT-XL/1
\citep{leng2025repae}; the random-label results are the re-evaluation
reported by \citet[Table~14]{zheng2025rae}.
VA-VAE uses LightningDiT-XL/1 at $800$ epochs
\citep[Table~3]{yao2025vavae}. RAE uses a frozen DINOv2-B encoder and
DiT$^{\mathrm{DH}}$-XL at $800$ epochs \citep{zheng2025rae}. Its latent
entry gives the spatial grid and channel count, $16\times16\times768$;
it is not a VAE with the compression settings of our models.
These comparisons evaluate complete systems, with different latent
dimensions and generator architectures.

\paragraph{DINO encoder size comparisons.}
REPA reports gFID $9.7$, $10.0$, and $9.8$ with DINOv2-B, L, and g
\citep[Table~2]{yu2025repa}; REPA-E reports $16.3$ and $16.0$ with B
and L \citep[Table~3]{leng2025repae}. These encoder-size studies are
distinct from Table~\ref{tab:vae_scale}, where REPA-E uses a fixed
DINO encoder and varies the diffusion backbone used for VAE tuning.

\begingroup
\paragraph{Text-to-image training and evaluation.}
\phantomsection
\label{app:t2i}
We use COCO 2014 at $256\times256$ resolution with the MMDiT
architecture from the REPA text-to-image setup \citep{yu2025repa}.
The model has $24$ blocks, hidden width $768$, and patch size $2$;
text conditioning uses frozen CLIP ViT-L/14 features.
All three VAEs use the $f8d4$ latent format.
We use the same tuned SD-VAE checkpoint as in Table~\ref{tab:vae_quality}(a),
obtained with SiT-XL and $\lambda_V=0.05$.
The ImageNet-tuned VAEs are transferred without further VAE tuning.
For each frozen VAE, we train a new MMDiT for $150$K updates
with global batch size $256$, using only the diffusion loss.

We evaluate EMA weights with the $250$-step Euler--Maruyama SDE
sampler, using CFG scale $1$ or $2$ for the unguided and guided
results, respectively. Each evaluation generates $40{,}192$ images
from the same COCO validation caption set. FID is computed against
the COCO validation reference statistics released with U-ViT
\citep{bao2023uvit}.
\par
\endgroup

\FloatBarrier
\section{Reconstruction Quality}
\label{app:reconstruction}
\begin{table}[!htbp]
\centering
\caption{\textbf{Generation--reconstruction trade-off.}
Ours uses $100$K tuning iterations with SiT-XL and an SD-VAE teacher.
gFID uses a new SiT-B trained for $400$K diffusion-only iterations on
each frozen VAE: B/2 for $f8d4$ and B/1 for $f16d32$.}
\label{tab:reconstruction}
\small
\setlength{\tabcolsep}{5pt}
\def\guareconrows{%
\toprule
VAE & $\lambda_V$ & PSNR$\uparrow$ & SSIM$\uparrow$ & LPIPS$\downarrow$ & rFID$\downarrow$ & gFID$\downarrow$\\
\midrule
\multicolumn{7}{l}{\textit{$f8d4$}}\\
SD-VAE & -- & 26.91 & 0.741 & 0.130 & 0.71 & 37.02\\
\rowcolor{oursorange}Ours & 0.05 & 25.38 & 0.716 & 0.136 & 0.75 & 22.53\\
\rowcolor{oursorange}Ours & 0.25 & 24.69 & 0.671 & 0.154 & 0.80 & 21.02\\
\midrule
\multicolumn{7}{l}{\textit{$f16d32$}}\\
IN-VAE & -- & 28.92 & 0.820 & 0.084 & 0.25 & 39.31\\
\rowcolor{oursorange}Ours & 0.05 & 27.70 & 0.792 & 0.097 & 0.34 & 19.30\\
\midrule
VA-VAE\dinomark & -- & 27.70 & 0.787 & 0.096 & 0.28 & 13.85\\
\bottomrule
}
\sbox{\guatablebox}{\begin{tabular}{lrrrrrr}\guareconrows\end{tabular}}
\setlength{\tabcolsep}{\dimexpr 5pt+(\linewidth-\wd\guatablebox)/14\relax}
\begin{tabular}{lrrrrrr}\guareconrows\end{tabular}
\end{table}

Table~\ref{tab:reconstruction} pairs reconstruction metrics with
downstream generation for both latent formats. Tuning improves
generation while degrading reconstruction relative to the original VAEs.
For SD-VAE, increasing $\lambda_V$ from
$0.05$ to $0.25$ further lowers gFID from $22.53$ to $21.02$, but reduces
PSNR from $25.38$ to $24.69$ and increases rFID from $0.75$ to $0.80$;
SSIM and LPIPS also worsen. We therefore use $0.05$ for the main XL
configuration as a compromise between generation and reconstruction,
while keeping $0.25$ fixed across sizes in the scaling experiment.

For IN-VAE, the selected $\lambda_V=0.05$ checkpoint lowers gFID from
$39.31$ to $19.30$, while PSNR decreases from $28.92$ to $27.70$ and
rFID increases from $0.25$ to $0.34$. Its PSNR, SSIM, and LPIPS are
close to VA-VAE's, although VA-VAE has lower rFID and better downstream
gFID. These results concern the tuned checkpoints themselves, without
additional decoder fine-tuning.

\FloatBarrier
\section{Representation Analysis}
\label{app:representations}
\paragraph{VAE checkpoints and visualization.}
The tuned SD-VAE and IN-VAE use SiT-XL with $\lambda_V=0.05$, matching
the selected VAEs in Table~\ref{tab:vae_quality}. PCA projects their
latent channels to three dimensions for qualitative inspection of
spatial patterns. Colors are not shared semantic labels across VAEs.

\paragraph{SRSS on VAE latents.}
Semantic-region self-similarity (SRSS) measures within-region structure
relative to foreground--background similarity \citep{singh2025irepa}.
We apply it to VAE latents following \citet{zhong2026diffusability}.
Images and foreground masks are resized to $256\times256$; posterior
modes give deterministic latents. Binary masks are max-pooled to each
latent grid, and images are retained when foreground and background
each occupy at least $25\%$ of tokens. We use the $1{,}024$ image--mask pairs provided with iREPA
\citep{singh2025irepa}. The foreground-size filter retains $469$
images for $f8d4$ and $517$ for $f16d32$. Latent channels are standardized over the retained images
and spatial locations. We use the same mask subset within each latent format and
the iREPA distance settings $d_{\mathrm{pos,min}}=d_{\mathrm{pos}}=1$
and $d_{\mathrm{neg}}=6$. Table~\ref{tab:representations} reports mean
SRSS, comparing each tuned VAE with its original counterpart and
REPA-E within the same format.

\paragraph{LNC on diffusion features.}
We adapt latent neighbor consistency \citep{zhong2026diffusability}
to diffusion features as a diagnostic of class-consistent neighborhoods.
Each SiT-B is trained from scratch for $400$K diffusion-only iterations
on its corresponding frozen VAE. We evaluate EMA weights at the output
of block $8$ and timestep $t=0.4$, before any projection or output head.
The null class label is used for feature extraction; ground-truth
labels are used only to evaluate neighbors. Thus the correct class
is not supplied to the diffusion model.

The evaluation uses the $12{,}419$ validation images of ImageNet-S
\citep{gao2022luss}, covering $919$ classes. Images and masks receive matched center crops at
$256\times256$. Latents are sampled from the VAE posterior and
normalized with the SiT checkpoint's fixed latent statistics before
adding noise. We use seed $0$, with matched noise draws across models
of the same latent shape. SiT-B/2 and SiT-B/1 both yield $16\times16$
token grids and $768$-dimensional features. Foreground masks are
resized to this grid by nearest-neighbor interpolation. Each feature
channel is standardized over images and spatial tokens, followed by
foreground mean pooling. The $72$ empty foreground masks yield zero
vectors and are retained consistently across all models. The SRSS
foreground-size filter is not applied to this evaluation.

For each pooled vector, we retrieve its ten nearest neighbors by
cosine distance, excluding itself, and average the fraction with the
same class label. Each class contributes at most $50$ images; no
class balancing is applied. Table~\ref{tab:representations} reports
LNC@10 as a percentage. The random-neighbor expectation for this set
is approximately $0.107\%$. LNC describes local semantic organization;
it is not a measurement of linear-probe accuracy. The small differences
between our method and REPA-E are single-seed observations and are not
interpreted as statistically significant.

\subsection{Latent Structure with a Shared Timestep}

\begin{figure}[!tbp]
\centering
\begin{minipage}{0.94\linewidth}%
\centering
\setlength{\lineskip}{0pt}%
\setlength{\parskip}{0pt}%
\edef\guapw{\the\dimexpr(\linewidth-12pt)/4\relax}%
\newcommand{\guasdvaeimage}[1]{\includegraphics[#1]{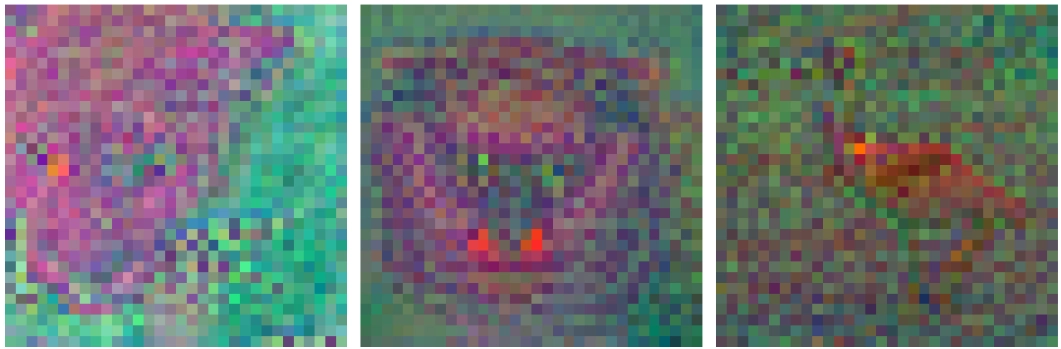}}%
\newcommand{\guaearlyimage}[1]{\includegraphics[#1]{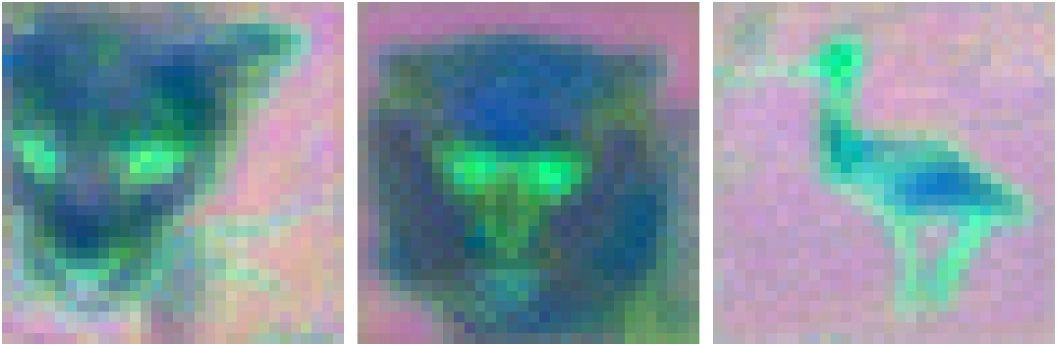}}%
\newcommand{\gualateimage}[1]{\includegraphics[#1]{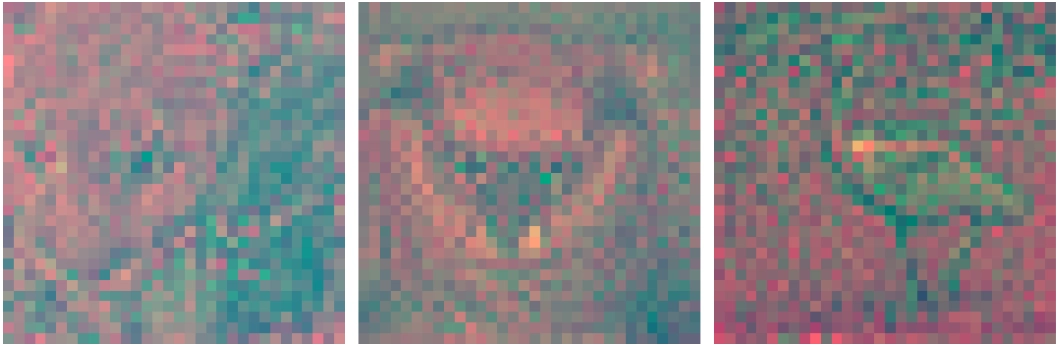}}%
\newcommand{\guapanel}[3]{%
  \sbox0{#3{}}%
  \edef\guatmp{\noexpand#3{viewport=%
    \the\dimexpr#1\wd0\relax\space\the\dimexpr0.009\ht0\relax\space
    \the\dimexpr#2\wd0\relax\space\the\dimexpr0.991\ht0\relax,%
    clip,width=\guapw,height=\guapw}}%
  \guatmp}%
\newcommand{\guarow}[3]{%
  \makebox[\linewidth]{%
  \includegraphics[viewport=#1,clip,width=\guapw,height=\guapw]{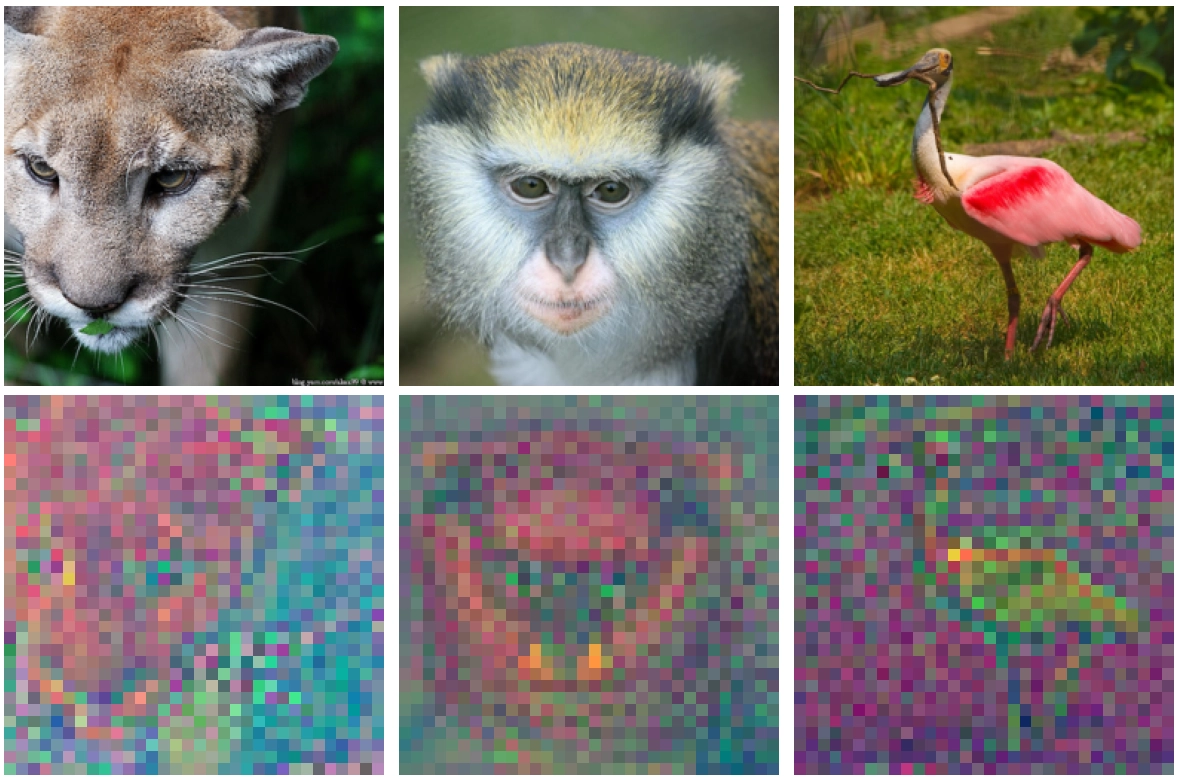}\hfill
  \guapanel{#2}{#3}{\guasdvaeimage}\hfill
  \guapanel{#2}{#3}{\guaearlyimage}\hfill
  \guapanel{#2}{#3}{\gualateimage}}\par}%
\newcommand{\guahead}[1]{\makebox[\guapw]{\small #1}}%
\makebox[\linewidth]{\guahead{Input}\hfill\guahead{SD-VAE}\hfill
\guahead{1,650 iterations}\hfill\guahead{100K iterations}}\par
\vspace{3pt}
\guarow{4 394 382 772}{0.0033}{0.3247}\vspace{4pt}
\guarow{399.5 394 777.5 772}{0.3393}{0.6607}\vspace{4pt}
\guarow{795 394 1173 772}{0.6753}{0.9967}\vspace{4pt}
\caption{\textbf{Evolution of latent structure with a shared timestep.}
Rows correspond to the same three input images. To the right of the inputs,
PCA visualizations show the original SD-VAE and two checkpoints of a
shared-timestep tuning run. The clearer structure at $1{,}650$ iterations
is not maintained at $100$K, although alignment remains active.
Colors are not shared semantic labels across columns.}
\label{fig:shared_timestep}
\end{minipage}
\end{figure}

\label{app:shared_timestep}
Figure~\ref{fig:shared_timestep} follows one preliminary SD-VAE tuning
run with shared teacher and student timesteps and CFG-enhanced teacher
features. Alignment remains active throughout training. Object structure
is more apparent at $1{,}650$ iterations, but becomes less coherent by
$100$K iterations, when the visualizations again resemble the original
SD-VAE. This qualitative observation motivates the timestep choice in
Section~\ref{sec:guidance}.

\section{Qualitative Results}
\label{app:qualitative}

Figure~\ref{fig:qualitative} shows samples from the 160-epoch $f16d32$
model in Table~\ref{tab:final_generation}. This model is obtained by
continuing the student jointly trained with IN-VAE during our VAE tuning stage,
using only the diffusion objective with the tuned VAE frozen. For visualization, we use a
50-step Euler ODE sampler and a CFG scale of 4.0 applied at all timesteps, instead of the
evaluation protocol in Appendix~\ref{app:evaluation}. For each of 48 ImageNet classes, we generate
eight samples and show one selected sample.

\begingroup
\makeatletter
\setlength{\@fptop}{0pt}
\makeatother
\begin{figure}[t]
  \centering
  \includegraphics[width=\linewidth]{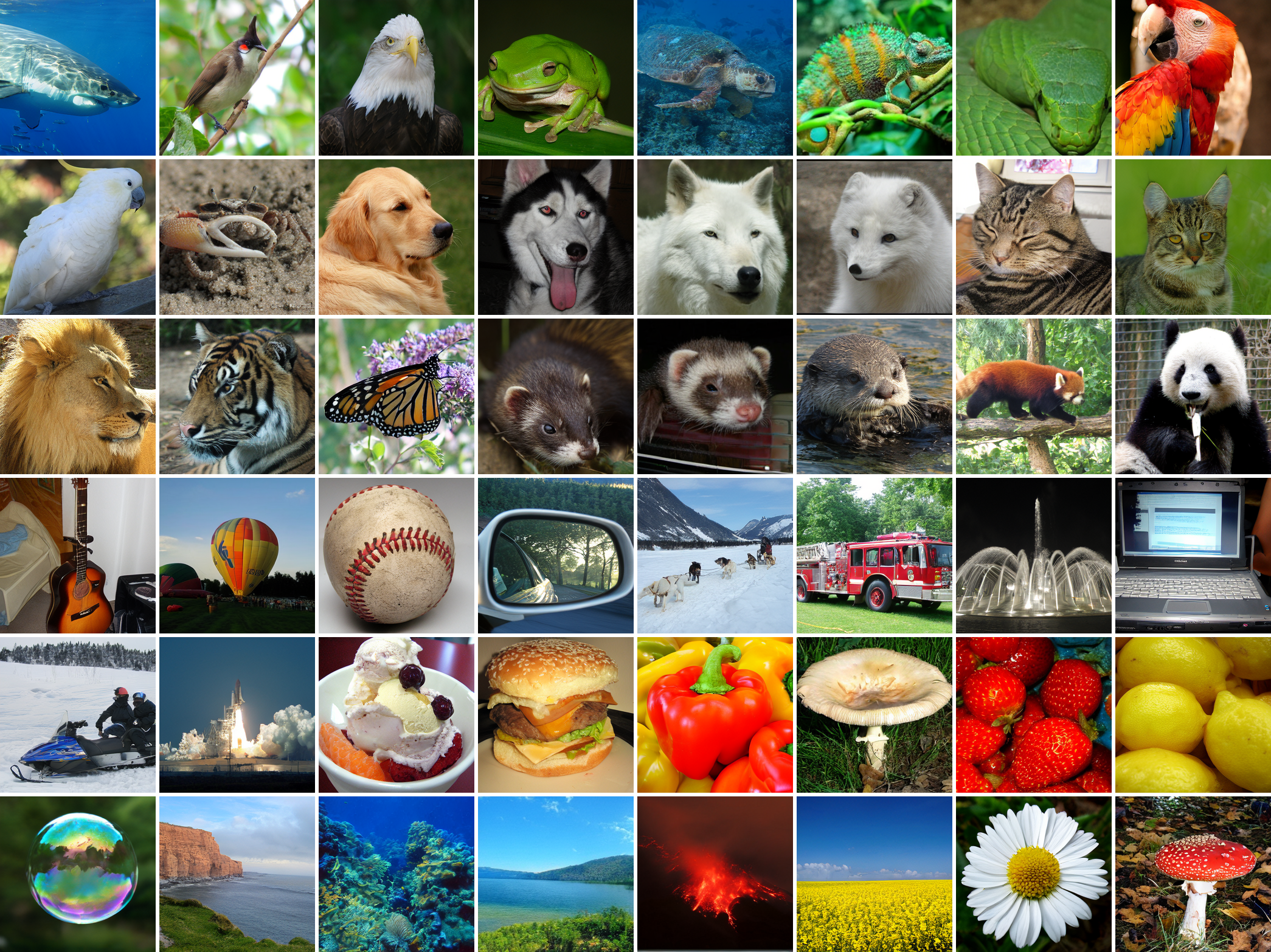}
  \caption{\textbf{Qualitative results on ImageNet $256\times256$ using our tuned VAE and SiT-XL/1.}
  We use a classifier-free guidance and show one selected sample per class.}
  \label{fig:qualitative}
\end{figure}
\clearpage
\endgroup

\end{document}